\pdfoutput=1

\documentclass[11pt]{article}

\usepackage{EMNLP2023}

\usepackage{times}
\usepackage{latexsym}

\usepackage[T1]{fontenc}

\usepackage[utf8]{inputenc}

\usepackage{microtype}

\usepackage{inconsolata}

\usepackage{xspace}
\usepackage{booktabs}
\usepackage{graphicx}
\usepackage{enumitem}
\usepackage{xcolor}
\usepackage{tikz-dependency}
\usepackage{url}
\usepackage{amssymb}
\usepackage{float}
\usepackage{todonotes}
\usepackage{csquotes}
\usepackage{xspace}
\usepackage{listings}
\usepackage{caption}
\usepackage{subcaption}
\usepackage[normalem]{ulem}

\setlist[description]{leftmargin=\parindent,labelindent=0pt}

\usepackage{tablefootnote}
\usepackage{tabularx}
\newcolumntype{L}{>{\raggedright\arraybackslash}X}
\usepackage{amsmath}
\usepackage{tikz}

\newcommand{\sref}[1]{Sec.~\ref{#1}}
\newcommand{\aref}[1]{Appendix~\ref{#1}}

\newcommand{\tref}[1]{Table~\ref{#1}}
\newcommand{\fref}[1]{Figure~\ref{#1}}

\renewcommand{\vec}[1]{\mathbf{#1}}

\newcommand{\corpusName}{Multi-Layer Materials Science Corpus\xspace}
\newcommand{\corpusNameShort}{MuLMS\xspace}

\newcommand{\neLabelMat}{\textsc{Mat}\xspace}
\newcommand{\neLabelNum}{\textsc{Num}\xspace}
\newcommand{\neLabelValue}{\textsc{Value}\xspace}
\newcommand{\neLabelUnit}{\textsc{Unit}\xspace}
\newcommand{\neLabelForm}{\textsc{Form}\xspace}
\newcommand{\neLabelProperty}{\textsc{Property}\xspace}
\newcommand{\neLabelCite}{\textsc{Cite}\xspace}
\newcommand{\neLabelSample}{\textsc{Sample}\xspace}
\newcommand{\neLabelTechnique}{\textsc{Technique}\xspace}
\newcommand{\neLabelRange}{\textsc{Range}\xspace}
\newcommand{\neLabelInstrument}{\textsc{Instrument}\xspace}
\newcommand{\neLabelDev}{\textsc{Device}\xspace}
\newcommand{\neLabelMeasurement}{\textsc{Measurement}\xspace}

\newcommand{\sofcNeLabelExperiment}{\textsc{Experiment}\xspace}
\newcommand{\sofcNeLabelMaterial}{\textsc{Material}\xspace}
\newcommand{\sofcNeLabelDevice}{\textsc{Device}\xspace}
\newcommand{\sofcNeLabelValue}{\textsc{Value}\xspace}

\newcommand{\msptNeLabelPropertyMisc}{\textsc{Property-Misc}\xspace}
\newcommand{\msptNeLabelPropertyUnit}{\textsc{Property-Unit}\xspace}
\newcommand{\msptNeLabelNumber}{\textsc{Number}\xspace}
\newcommand{\msptNeLabelCharacterizationApparatus}{\textsc{Characterization-Apparatus}\xspace}
\newcommand{\msptNeLabelApparatusUnit}{\textsc{Apparatus-Unit}\xspace}
\newcommand{\msptNeLabelConditionMisc}{\textsc{Condition-Misc}\xspace}
\newcommand{\msptNeLabelMeta}{\textsc{Meta}\xspace}
\newcommand{\msptNeLabelSynthesisApparatus}{\textsc{Synthesis-Apparatus}\xspace}
\newcommand{\msptNeLabelOperation}{\textsc{Operation}\xspace}
\newcommand{\msptNeLabelAmountMisc}{\textsc{Amount-Misc}\xspace}
\newcommand{\msptNeLabelAmountUnit}{\textsc{Amount-Unit}\xspace}
\newcommand{\msptNeLabelReference}{\textsc{Reference}\xspace}
\newcommand{\msptNeLabelPropertyType}{\textsc{Property-Type}\xspace}
\newcommand{\msptNeLabelMaterial}{\textsc{Material}\xspace}
\newcommand{\msptNeLabelMaterialDescriptor}{\textsc{Material-Descriptor}\xspace}
\newcommand{\msptNeLabelApparatusDescriptor}{\textsc{Apparatus-Descriptor}\xspace}
\newcommand{\msptNeLabelApparatusPropertyType}{\textsc{Apparatus-Property-Type}\xspace}
\newcommand{\msptNeLabelConditionUnit}{\textsc{Condition-Unit}\xspace}
\newcommand{\msptNeLabelNonrecipeMaterial}{\textsc{Nonrecipe-Material}\xspace}
\newcommand{\msptNeLabelConditionType}{\textsc{Condition-Type}\xspace}
\newcommand{\msptNeLabelBrand}{\textsc{Brand}\xspace}

\newcommand{\relLabelHasForm}{\textit{hasForm}\xspace}
\newcommand{\relLabelMeasuresProperty}{\textit{measures\-Property}\xspace}
\newcommand{\relLabelPropertyValue}{\textit{property\-Value}\xspace}
\newcommand{\relLabelUsedAs}{\textit{usedAs}\xspace}
\newcommand{\relLabelConditionProperty}{\textit{conditionProperty}\xspace}
\newcommand{\relLabelConditionSample}{\textit{conditionSample}\xspace}
\newcommand{\relLabelUsesTechnique}{\textit{usesTechnique}\xspace}
\newcommand{\relLabelUsedTogether}{\textit{usedTogether}\xspace}
\newcommand{\relLabelConditionEnvironment}{\textit{conditionEnvironment}\xspace}
\newcommand{\relLabelUsedIn}{\textit{usedIn}\xspace}

\newcommand{\relLabelTakenFrom}{\textit{takenFrom}\xspace}
\newcommand{\relLabelDopedBy}{\textit{dopedBy}\xspace}
\newcommand{\relLabelMeasuresPropertyValue}{\textit{measuresPropertyValue}\xspace}
\newcommand{\relLabelConditionPropertyValue}{\textit{conditionPropertyValue}\xspace}

\newcommand{\sofcRelLabelAnodeMaterial}{\textit{AnodeMaterial}\xspace}
\newcommand{\sofcRelLabelCathodeMaterial}{\textit{CathodeMaterial}\xspace}
\newcommand{\sofcRelLabelConductivity}{\textit{Conductivity}\xspace}
\newcommand{\sofcRelLabelCurrentDensity}{\textit{CurrentDensity}\xspace}
\newcommand{\sofcRelLabelDegradationRate}{\textit{DegradationRate}\xspace}
\newcommand{\sofcRelLabelDevice}{\textit{Device}\xspace}
\newcommand{\sofcRelLabelElectrolyteMaterial}{\textit{ElectrolyteMaterial}\xspace}
\newcommand{\sofcRelLabelFuelUsed}{\textit{FuelUsed}\xspace}
\newcommand{\sofcRelLabelInterlayerMaterial}{\textit{InterlayerMaterial}\xspace}
\newcommand{\sofcRelLabelOpenCircuitVoltage}{\textit{OpenCircuitVoltage}\xspace}
\newcommand{\sofcRelLabelPowerDensity}{\textit{PowerDensity}\xspace}
\newcommand{\sofcRelLabelResistance}{\textit{Resistance}\xspace}
\newcommand{\sofcRelLabelSupportMaterial}{\textit{SupportMaterial}\xspace}
\newcommand{\sofcRelLabelTimeOfOperation}{\textit{TimeOfOperation}\xspace}
\newcommand{\sofcRelLabelVoltage}{\textit{Voltage}\xspace}
\newcommand{\sofcRelLabelWorkingTemperature}{\textit{WorkingTemperature}\xspace}

\newcommand{\msptRelLabelRecipeTarget}{\textit{Recipe-target}\xspace}
\newcommand{\msptRelLabelSolventMaterial}{\textit{Solvent-material}\xspace}
\newcommand{\msptRelLabelAtmosphericMaterial}{\textit{Atmospheric-material}\xspace}
\newcommand{\msptRelLabelRecipePrecursor}{\textit{Recipe-precursor}\xspace}
\newcommand{\msptRelLabelParticipantMaterial}{\textit{Participant-material}\xspace}
\newcommand{\msptRelLabelApparatusOf}{\textit{Apparatus-of}\xspace}
\newcommand{\msptRelLabelConditionOf}{\textit{Condition-of}\xspace}
\newcommand{\msptRelLabelDescriptorOf}{\textit{Descriptor-of}\xspace}
\newcommand{\msptRelLabelNumberOf}{\textit{Number-of}\xspace}
\newcommand{\msptRelLabelAmountOf}{\textit{Amount-of}\xspace}
\newcommand{\msptRelLabelApparatusAttrOf}{\textit{Apparatus-attr-of}\xspace}
\newcommand{\msptRelLabelBrandOf}{\textit{Brand-of}\xspace}
\newcommand{\msptRelLabelCoreOf}{\textit{Core-of}\xspace}
\newcommand{\msptRelLabelNextOperation}{\textit{Next-operation}\xspace}

\newcommand{\meas}{\textsc{Measurement}\xspace}
\newcommand{\qualmeas}{\textsc{Qual\_Meas}\xspace}
\newcommand{\None}{\textsc{None}\xspace}

\newcommand{\threerows}[3]{\begin{tabular}{@{}c@{}c@{}}#1 \\ #2 \\ #3\end{tabular}}

\newcommand{\totalNumSentences}{10186\xspace} %
\newcommand{\totalNumEntityAnnots}{46,351\xspace}

\makeatletter
\def\url@leostyle{%
  \@ifundefined{selectfont}{\def\UrlFont{\sf}}{\def\UrlFont{\scriptsize\sffamily}}}
\makeatother
\title{\corpusNameShort: A Multi-Layer Annotated Text Corpus\\ for Information Extraction in the Materials Science Domain}

\author{Timo Pierre Schrader$^{1,3}$~
  Matteo Finco$^2$~
  Stefan Grünewald$^{1,4}$\\
  {\bf Felix Hildebrand$^2$}~
  {\bf Annemarie Friedrich$^3$} \\
  $^1$Bosch Center for Artificial Intelligence, Renningen, Germany \\ 
    $^2$Robert Bosch GmbH, Stuttgart, Germany \hspace{-1.2mm} \\ $^3$University of Augsburg, Germany \hspace{2.0mm} $^4$University of Stuttgart, Germany \\
\texttt{timo.schrader|matteo.finco|stefan.gruenewald@de.bosch.com} \\
  \texttt{annemarie.friedrich@informatik.uni-augsburg.de}}

\date{}

\begin{document}
\maketitle
\begin{abstract}
Keeping track of all relevant recent publications and experimental results for a research area is a challenging task.
Prior work has demonstrated the efficacy of information extraction models in various scientific areas.
Recently, several datasets have been released for the yet understudied materials science domain.
However, these datasets focus on sub-problems such as parsing synthesis procedures or on sub-domains, e.g., solid oxide fuel cells.

In this resource paper, we present \corpusNameShort, a new dataset of 50 open-access articles, spanning seven sub-domains of materials science.
The corpus has been annotated by domain experts with several layers ranging from named entities over relations to frame structures.
We present competitive neural models for all tasks and demonstrate that multi-task training with existing related resources leads to benefits.
\end{abstract}

\section{Introduction}
\label{sec:intro}
Designing meaningful experiments in empirical sciences requires maintaining a detailed overview of the huge amounts of literature published every year.
Applying natural language processing (NLP) in this context has risen to be an active research area \citep{sdp-2020-scholarly,sdp-2021-scholarly,sdp-2022-scholarly}.
Besides the biomedical field, which has been studied extensively in the past decades \citep[e.g.,][]{bionlp2004,ws-2012-bionlp,bionlp-2022-biomedical}, the less-studied materials science domain has recently received more attention \citep{mysore-etal-2019-materials,friedrich-etal-2020-sofc,ogorman-etal-2021-ms}.

Materials science research aims to design and discover new materials.
Part of the papers is hence often dedicated to %
the \textit{synthesis procedures}, %
the \enquote{recipe} for creating a material.
Their extraction from papers
has been covered by \citet{mysore-etal-2019-materials} and \citet{ogorman-etal-2021-ms}.
Much materials science research develops materials in the context of creating a particular device, e.g., batteries or photovoltaic panels. %
The device is tested in various conditions and the literature needs to be analysed for identifying promising set-ups.
\citet{friedrich-etal-2020-sofc} address this for solid oxide fuel cells.

\begin{figure*}[t]
\centering
\includegraphics[width=1.0\textwidth]{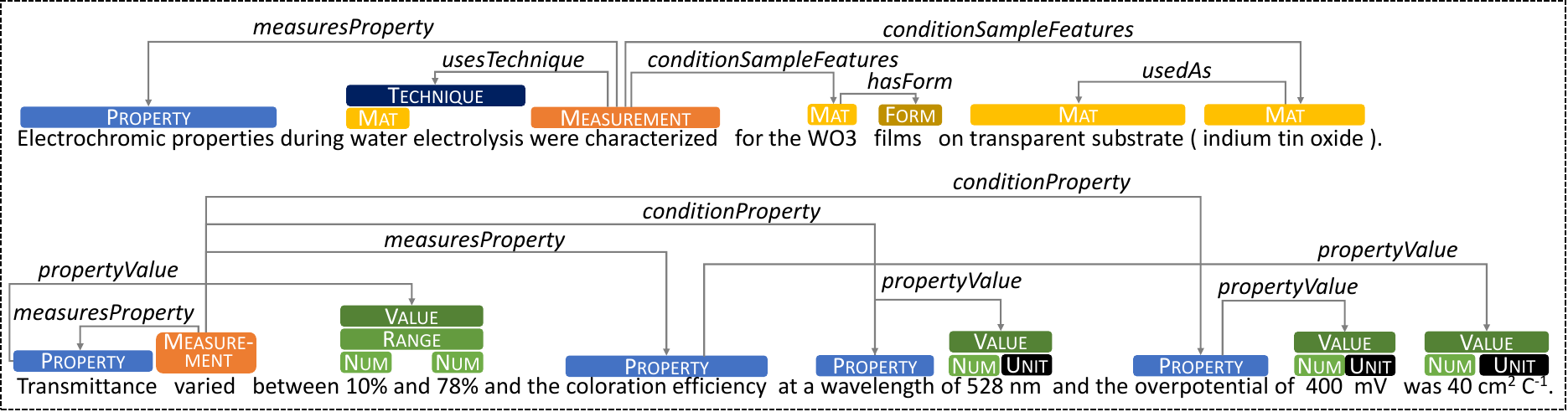}
	\caption{\textbf{\corpusName}: named entity, relation and semantic role annotations.}
	\label{fig:teaser}
\end{figure*}

In this paper, we introduce \textbf{\corpusNameShort} (the \underline{Mu}lti-\underline{L}ayer \underline{M}aterials \underline{S}cience corpus), a new dataset of scientific publications annotated by domain experts with named entity (NE) mentions, relations, and frame structures corresponding to a broad notion of measurements (see \fref{fig:teaser}).
In contrast to prior works, %
we include papers from a variety of materials science sub-domains.
To the best of our knowledge, the existing datasets only annotate particular paragraphs or subsets of the sentences with NE mentions.
Our dataset is the first to \textit{exhaustively} annotate a large-scale collection of materials science articles with NEs and facilitates novel semantic search applications, e.g., answering search queries such as \enquote{find a passage within a paper reporting a measurement using material X, condition Y, and obtaining a value of at least Z.}

The design of \corpusNameShort' annotation scheme results from a collaboration of NLP and materials science experts.
Our inter-annotator agreement study shows good agreement for most categories and decisions.
We propose several machine learning tasks on the annotated data and present strong neural baselines for all tasks, which signals a high level of consistency across the annotations in our dataset.
We cast detecting sentences describing measurements as a sentence classification task and provide a robust tagger for recognizing NEs.
We propose to treat relation and semantic role extraction on \corpusNameShort in a single step using a dependency parser that predicts relations between the NEs in a sentence.
According to our multi-task experiments with related datasets, training jointly with \corpusNameShort is beneficial for performance on those datasets.

Our contributions are as follows. %
    (1) We publicly release a dataset of 50 open-access scientific publications \textit{exhaustively} annotated by a domain expert with NE mentions, relations, and measurement frames.\footnote{\url{https://github.com/boschresearch/mulms-wiesp2023}} %
    (2) We define a set of NLP tasks on \corpusNameShort and provide strong transformer-based baselines. Our code will be published. %
    (3) We formulate the relation and frame-argument extraction as a single 
    dependency parsing task, which extracts \textit{all} relations in a sentence in one processing step.
    (4) We perform an extensive set of multi-task learning experiments with related corpora, showing that \corpusNameShort is a useful auxiliary task for two other materials science NLP datasets.

\section{Related Work}
\label{sec:relwork}
Several \textbf{materials science NLP datasets} have recently been released, e.g., targeting NE recognition \citep{yamaguchi-etal-2020-sc,ogorman-etal-2021-ms}.
The Materials Science Procedural Text (MSPT) corpus \citep{mysore-etal-2019-materials} consists of paragraphs describing synthesis procedures annotated with graph structures capturing relations and typed arguments.
SOFC-Exp \citep{friedrich-etal-2020-sofc} marks similar graph structures describing experiments. %

In this paper, we compare two state-of-the-art approaches to \textbf{Named Entity Recognition} (NER).
\citet{huang2015} use a CRF layer \citep{lafferty2001CRF} on top of a neural language model (in their case a BiLSTM) for sequence-tagging related tasks.
\citet{yu-etal-2020-named} treat NER as a graph-based dependency parsing task
by representing NEs as spans between the first and last token of an entity. %
In the materials science domain, \citet{friedrich-etal-2020-sofc} test a variety of embedding combinations in a CRF-based tagger.
\citet{ogorman-etal-2021-ms} compare different pre-trained transformers for token classification. %
Both studies find SciBERT \citep{beltagy-etal-2019-scibert}, a BERT-style model pre-trained on scientific documents, to be very effective. %

\textbf{Relation and Event Extraction.}
\citet{friedrich-etal-2020-sofc} treat their slot filling task in the SOFC sub-domain as a sequence tagging task, assuming that each sentence represents at most one experiment.
To predict a possible relation between two entities, \citet{swarup-etal-2020-instance} retrieve a set of sentences similar to the input sentence, and learn to copy relations from these sentences.
\citet{mysore2017automatically} experiment with unsupervised methods for extracting action graphs for synthesis procedures.

An exhaustive overview of the literature on biomedical relation extraction is out of the scope of this paper.
Recent works have used graph-neural networks \citep{huang-etal-2020-biomedical}, or convolutional neural networks \citep{ramponi-etal-2020-cross}.
\citet{sarrouti-etal-2022-comparing} compare various pre-trained transformer models.
Semantic parsing of frame structures \citep{fillmore2001frame} has been addressed using graph-convolutional networks \citep{marcheggiani-titov-2020-graph}, BiLSTMs \citep{he-etal-2018-jointly}, and recently by generating structured output using encoder-decoder models \citep{hsu-etal-2022-degree,lu-etal-2021-text2event}.
Tackling semantic dependency parsing with a biaffine classifier architecture was first proposed by \citet{dozat-manning-2018-simpler}.

\section{\corpusNameShort Corpus}
\label{sec:data}
In this section, we present our new corpus.

\subsection{Source of Texts and Preprocessing}
We select 50 scientific articles licensed under CC BY from seven popular sub-areas of materials science: electrolysis, graphene, polymer electrolyte fuel cell (PEMFC), solid oxide fuel cell (SOFC), polymers, semiconductors, and steel.
The four SOFC papers were selected from the SOFC-Exp corpus \cite{friedrich-etal-2020-sofc}.
11 papers were selected from the OA-STM corpus\footnote{\href{https://github.com/elsevierlabs/OA-STM-Corpus}{https://github.com/elsevierlabs/OA-STM-Corpus}} and classified into the above subject areas by a domain expert.
The majority of the papers were found via PubMed\footnote{\href{https://pubmed.ncbi.nlm.nih.gov/}{https://pubmed.ncbi.nlm.nih.gov}} and DOAJ\footnote{\href{https://doaj.org}{https://doaj.org}} using queries prepared by a domain expert.
For the OA-STM data, we use the sentence segmentation provided with the corpus, which has been created using GENIA tools \citep{tsuruoka-tsujii-2005-bidirectional}.
For the remaining texts, we rely on the sentence segmentation provided by INCEpTION \citep{klie-etal-2018-inception} with some manual fixes.
As shown in \tref{tab:basic_stats}, documents are rather long with a tendency to long sentences (but with high variation due to, i.a., short headings).

\subsection{Annotation Scheme}
We annotate various layers: NEs, relations, and frame structures representing measurements.

\subsubsection{Named Entities}
\label{sec:ne_labels}
We annotate the following materials-science specific NE mentions and assign the following NE types to these mentions:\vspace{-3mm}

\begin{description}
    \setlength{\itemsep}{0pt}
    \setlength{\parskip}{0pt}
    \setlength{\parsep}{0pt}
\item[\textsc{Mat}:] mentions of materials described by their chemical formula (\textit{WO$_3$}) or its chemical name (\textit{indium tin oxide}).
\item[\textsc{Form}:] mentions of the form or morphology of the material, e.g., \textit{thin film}, \textit{gas}, \textit{liquid}, \textit{cubic}.%
\item[\textsc{Instrument}:] mentions of devices used to perform a materials-science-related measurement, e.g., \textit{Olympus BX52 microscope}.
\item[\textsc{Device}:] mentions of devices (target products) whose construction or improvement is the aim of the research (e.g., \textit{photodetector}, \textit{transistor}, \textit{supercapacitor}). \textsc{Device} is not used for instrumentation devices that are only used as a tool.
\item[\textsc{Num}:] mentions of numbers such as \textit{0.46}.
\item[\textsc{Unit}:] mentions of units such as \textit{nm} or \textit{V}.
\item[\textsc{Range}:] mentions of numeric expressions indicating ranges, e.g., \textit{0.46}$\pm$\textit{.11}
\item[\textsc{Value}:] nested type capturing expressions of values usually composed of a \textsc{Num}, \textsc{Range} and a \textsc{Unit}, e.g., \char`\~ \textit{5 x 3mm2}
\item[\textsc{Cite}:] citations, e.g., \textit{Setman et al.} or \textit{[13]}.
\item[\textsc{Property}:] expressions referring to properties of materials or conditions in experiments, e.g., \textit{stress rate} or \textit{electron conductivity}.
\item[\textsc{Technique}:] mentions of experimental techniques used in the characterization steps, e.g., \textit{Scanning electron microscopy (SEM)}.
\item[\textsc{Sample}:] mention of the material or a component made of materials studied in a measurement, either referred to by a particular name or its composition, its batch name (\textit{Aq-825}) or by referring to the whole component (\textit{MEA-Pt/C}) or to part of the material's structure (\textit{ionomer patches}).
In simulation papers, the \textsc{Sample} may also be the computational model under study (\textit{RBF-ANN}).
\end{description}
\vspace{-1mm}

\begin{table}[t]
\footnotesize
\centering
\setlength\tabcolsep{8pt}
\begin{tabular}{lr}
\toprule
     \#Documents & 50\\
     \#Documents train / dev / test & 36 / 7 / 7\\
    \#Sentences & \totalNumSentences\\ %
    \#Sentences/Document & 203.7$\pm$73.2\\ %
    \#Tokens/Sentence & 28.7$\pm$17.9\\ %
     \bottomrule
\end{tabular}
\caption{Basic \textbf{corpus statistics} for \corpusNameShort.}
\label{tab:basic_stats}
\end{table}

\subsubsection{Relations and Measurement Frame}
We treat measurement annotation in a frame-like \citep{fillmore2001frame} fashion, using the span type \meas to mark the triggers (e.g., \textit{was measured}, \textit{is plotted}) that introduce the \textit{Measurement} frame to the discourse.
About 88\% of the triggers are verbs.
The remaining 12\% occur in figure captions without verb phrases and are annotated either on nouns (\textit{Comparison}) or, in absence of more suitable phrases, on figure labels such as \textit{Figure 17}.
The trigger annotations of these sentences serve as the root of the tree/graph annotations as illustrated in \fref{fig:teaser}.

There are also cases in which the \textit{Measurement} frame is evoked, but there are no technical details or results that we can extract about the measurement.
We mark the triggers of these sentences with the tag \qualmeas (qualitative mention of a measurement).
An example of such a sentence is \enquote{\textit{We compare a critical volume to be detached from the different nanostructures.}} \vspace{1mm}

\textbf{Measurement-related Relations.}
We annotate several relations that start at a \textsc{Measurement} tag and that end at the annotations of the corresponding slot fillers within the sentence.
Consider the following sentence: \enquote{\textit{To characterize the ORR activity of the catalyst, linear scan voltammetry (LSV) was tested from 0 to 1.2 V on an RDE with a scan rate of 50 mV/s in O2-saturated HClO4 solution.}}\vspace{-2mm}

\begin{description}
    \setlength{\itemsep}{0pt}
    \setlength{\parskip}{0pt}
    \setlength{\parsep}{0pt}
\item[\textit{measuresProperty}:] indicates the \textsc{Property} (e.g., \textit{ORR activity}) that is measured.
\item[\textit{conditionSampleFeatures}:] indicates the \textsc{Sample} or \textsc{Material} whose property is measured. In the above example, the sample is the \textit{catalyst}. %
\item[\textit{usesTechnique}:] relates to the \textsc{Technique} (e.g., \textit{linear scan voltammetry}) used in a measurement.
\item[\textit{conditionInstrument}:] refers to the \textsc{Instrument} used to make a measurement, e.g., \textit{RDE}/\textit{rotating disk electrode}.
\item[\textit{conditionProperty}:] a property that is a condition in the experiment, e.g., \textit{scan rate} (which in turn has the \textit{propertyValue} of \textit{50 mv/s}).
\item[\textit{propertyValue}:] connects the mention of a \textsc{Property} and that of its corresponding \textsc{Value}. This relation may also occur if a mention of a \textsc{Property} occurs independently of a measurement.
\item[\textit{conditionEnvironment}:] identifies the \textsc{Material}s (e.g., \textit{O2} and \textit{HClO4}) and \textsc{Value}s (e.g., an operating temperature of \textit{30°C}) that provide the environment of the measurement.
\item[\textit{takenFrom}:] connects the \textsc{Measurement} with the bibliographic reference \textsc{Cite} from which the setup has been inspired or taken over.
\end{description}

In most cases, a \relLabelConditionProperty or a \relLabelMeasuresProperty connects the \meas annotation to a \neLabelProperty node, at which a \relLabelPropertyValue relation starts that ends at the respective \neLabelValue.
However, in some cases, the condition or measured property is not mentioned explicitly.
In this case, we link the \neLabelValue directly to the \meas node via a \relLabelConditionPropertyValue or a \relLabelMeasuresPropertyValue link.
For consistency reasons, we also add these links in cases that mention the property explicitly, turning the trees into graph structures.
Out of the added \relLabelConditionPropertyValue links, 967 are for such explicit cases, while the other 206 describe implicit cases. In the case of \relLabelMeasuresPropertyValue, 722 links are for explicit cases and 36 for implicit cases.

\begin{table}[t]
\footnotesize
\centering
\setlength\tabcolsep{4pt}
\begin{tabular}{llr}
\toprule
     \textbf{Relation} & \textbf{Example} \\
     \midrule
     \textit{hasForm} & silicon\textsubscript{\textsc{Mat}}--\textit{hasForm}$\rightarrow$ hexagonal\textsubscript{\textsc{Form}}\\ %
     \textit{usedIn} & Sic\textsubscript{\textsc{Mat}} --\textit{usedIn}$\rightarrow$ MOSFET\textsubscript{\textsc{Device}} \\%& 446 \\
     \textit{usedAs} & PtNi3M\textsubscript{\textsc{Mat}} --\textit{usedAs}$\rightarrow$ catalysts \\ %
     \textit{dopedBy} & chlorinated\textsubscript{Material}--\textit{dopedBy}$\rightarrow$SiC\textsubscript{\textsc{Mat}} \\ %
     \bottomrule
\end{tabular}
\caption{Measurement-independent \textbf{relations} annotated in \corpusNameShort. \textsc{Mat} is short for \textsc{Material}.}
\label{tab:relation-overview}
\end{table}

\begin{table}[t]
\footnotesize
\centering
\setlength\tabcolsep{2pt}
\begin{tabular}{lrr|lrr}
\toprule
\textbf{Label} &   \textbf{Count} & \textbf{\%}  &  \textbf{Label} &   \textbf{Count} & \textbf{\%} \\
\midrule
\textsc{Mat} &   15596  & 33.6 & \textsc{Cite} &    1709  & 3.7\\
\textsc{Num} &    6081  &  13.1 & \textsc{Sample} &    1461  &  3.2 \\
\textsc{Value} &    4852  & 10.5 & \textsc{Technique} &    1036 & 2.2 \\
\textsc{Unit} &    4330  & 9.3 &  \textsc{Dev} &     808  &  1.7\\
\textsc{Property} &    3925  & 8.5 &  \textsc{Range} &     736 & 1.6\\
\textsc{Form} &    3568  & 7.7 & \textsc{Instrument} &     378 & 0.8\\
\textsc{Measurement} &    2171  & 4.7 & \textit{total} & \totalNumEntityAnnots& -\\
\bottomrule
\end{tabular}
\caption{Corpus counts for \textbf{named entity} annotations.}
\label{tab:ner-counts-condensed}
\end{table}

\paragraph{Further Relations.}
In the following, we explain relations that can appear independently of measurements.
Examples are shown in \tref{tab:relation-overview}.

\begin{description}
    \setlength{\itemsep}{0pt}
    \setlength{\parskip}{0pt}
    \setlength{\parsep}{0pt}
    \item[\textit{hasForm}:] connects mention of \textsc{Material} and the corresponding \textsc{Form} annotation.
    \item[\textit{usedIn}:] connects \textsc{Material} and the \textsc{Device} it is used in. %
    In \tref{tab:relation-overview}, \textit{MOSFET} stands for \textit{Metal Oxide Semiconductor Field-Effect Transistors}.
    \item[\textit{usedAs}:] links a specific \textsc{Material} mention with a more generic one such as \textit{catalyst}, a material class defined by its function. %
    \item[\textit{dopedBy}:] indicates dopants (e.g., \textit{chlorine}), i.e., impurities added to a main material (e.g., \textit{SiC}). %
    \item[\textit{usedTogether}:] connects two \textsc{Material}s if they are used together in an experiment, i.e., if the materials are part of an assembly or a mixture. %
\end{description}

\subsection{Corpus Statistics}
We now analyze our corpus and provide detailed corpus statistics.
In total, there are \totalNumEntityAnnots NE annotations.
\tref{tab:ner-counts-condensed} shows the counts by NE label.
There are roughly 1.5 MAT annotations per sentence as these are nested and occurrences of composite materials often result in many combined MAT tags.
\tref{tab:relCounts-condensed} reports the counts of annotated relations (16,794 in total), with \relLabelHasForm as the most frequent relation with 2910 instances and \relLabelDopedBy the least frequent with only 65 instances.

\begin{table}[t]
\centering
\footnotesize
\setlength\tabcolsep{2pt}
\begin{tabular}{lrr|lrr}
\toprule
                  \textbf{Label} & \textbf{Count} & \textbf{\%} & \textbf{Label} & \textbf{Count} & \textbf{\%}\\
\midrule
                \relLabelHasForm &   2910 & 17.3 & \textit{meas.Prop.Val.} & 751 & 4.5\\
       \relLabelMeasuresProperty &   2080 & 12.4 & \relLabelUsedTogether & 672 & 4.0\\
                 \relLabelUsedAs &   1839 &  11.0 &  \textit{conditionEnv.} &    549 & 3.3\\
          \relLabelPropertyValue &   1794 & 10.7 & \relLabelUsedIn &    434  & 2.6\\
      \relLabelConditionProperty &   1648 & 9.8 &\textit{conditionInstr.} &    357  & 2.2 \\
\relLabelConditionSample &   1434 & 8.5 & \relLabelTakenFrom &    118 & 0.7\\
\textit{cond.Prop.Value} &   1158 & 6.9 & \relLabelDopedBy &     65 & 0.4\\
         \relLabelUsesTechnique &    985 & 5.9 & \textit{total} & \multicolumn{2}{c}{16,794}\\
\bottomrule
\end{tabular}
\caption{Corpus counts for \textbf{measurement relations}.} %
\label{tab:relCounts-condensed}
\end{table}

\begin{table}[t]
    \centering
    \footnotesize
\begin{tabular}{lrrr}
\toprule
{} &  MEAS &  QUAL\_MEAS &  OTHER \\
\midrule
MEAS      &    48 &          6 &      6 \\
QUAL\_MEAS &    12 &         37 &     11 \\
OTHER     &    10 &         17 &     92 \\
\bottomrule
\end{tabular}
    \caption{Inter-annotator agreement for \textbf{identifying measurement sentences}: confusion matrix.}
    \label{tab:agreement-study-measurement-sentences}
\end{table}

Out of all \totalNumSentences sentences, 2111 (20.7\%) describe a measurement (i.e., they contain at least one \meas annotation).
On average, each document contains 43.4 \meas annotations.
In addition, there are 1476 sentences (14.5\%) marked as containing a \qualmeas, with 40 sentences of these also containing a \meas annotation. %

\subsection{Inter-Annotator Agreement (IAA)}
\label{sec:iaa-discussion}
Our entire dataset has been annotated by a graduate student of materials science, who was also involved in the design of the annotation scheme.
We perform two agreement studies, comparing to the annotations of a second annotator with a PhD degree in environment engineering and several years of experience in materials science research.

\textbf{Agreement on identifying Measurement sentences.}
In this agreement study, we estimate the degree of agreement whether a sentence expresses a \meas, %
a \qualmeas, or whether it does not express a measurement at all.
We sample 60 sentences marked with \meas, 60 sentences marked with \qualmeas, and 120 sentences not marked as either by the first annotator. %
\tref{tab:agreement-study-measurement-sentences} shows the confusion matrix for the 239 sentences for which both annotators provided a label.
One automatically selected sentence was not labeled by one of the annotators due to incomprehensibility.
In terms of Cohen's $\kappa$ \citep{cohen1960kappa}, agreement amounts to 59.2, indicating moderate to substantial agreement \citep{landis1977application}.
When collapsing \meas and \qualmeas, $\kappa$ is 63.4 (substantial).

\begin{table}[t]
    \centering
    \footnotesize
    \setlength\tabcolsep{1.5pt}
\begin{tabular}{lrrr|lrrr}
\toprule
      \textbf{Label} &  \textbf{P} &  \textbf{R} &  \textbf{F1} & \textbf{Label} &  \textbf{P} &  \textbf{R} & \textbf{F1}\\
\midrule
        \neLabelMat &       96.7 &    91.2 &   93.9     &    \neLabelCite &       97.4 &    97.4 & 97.4\\
        \neLabelNum &       98.9 &   100.0 &  99.4       &    \neLabelSample &        3.1 &    12.5 & 4.7\\
      \neLabelValue &       100 &    100 &  100.0        &    \textsc{Techn.} &       77.5 &    59.6 & 67.4 \\
       \neLabelUnit &       97.9 &   100.0 &  98.9       &   \neLabelDev &       96.0 &    82.8 & 88.9\\
   \neLabelProperty &       42.6 &    37.7 &  40.0       &  \neLabelRange &       100.0 &    100.0 & 100.0\\
       \neLabelForm &       95.7 &    86.3 &  90.8     & \textsc{Instr.} &       80.0 &    76.9 & 78.4\\
\textsc{Meas.} &       44.6 &    51.0 & 47.6      &  \textit{average} & \textit{79.3} & \textit{76.6} & \textit{77.9}\\
\bottomrule
\end{tabular}
    \caption{Inter-annotator agreement: \textbf{named entities}.}
    \label{tab:iaa-named-entities}
\end{table}

\textbf{Agreement on named entities.}
We next compute agreement for NE and relation annotations. %
IAA on \enquote{easy} types such as \textsc{Mat}, \textsc{Num}, \textsc{Unit}, \textsc{Value} and \textsc{Range} has been shown to be very high in prior work \citep{friedrich-etal-2020-sofc}.
Hence, as our resources are limited, we provide annotations of these types for correction to the second annotator.
We sample 134 sentences such that each entity type occurs at least 25 times in the annotations of the first annotator and have the second annotator correct or add entity annotations.
We then compare the annotated sets of NE mentions using precision and recall (for a justification of this choice of agreement metrics, see \aref{sec:app-detailed-corpus-stats}).
Results using relaxed matching (containment) are shown in \tref{tab:iaa-named-entities} (detailed counts in \aref{sec:app-detailed-corpus-stats}).
For most types, scores are in the expected range of difficult semantic annotation tasks.
Agreement on identifying Measurement sentences %
is good; the decision of where exactly to place the \meas annotation differs between annotators.

\begin{table}[t]
    \centering
    \footnotesize
    \setlength\tabcolsep{5pt}
\begin{tabular}{lrrr|rr}
\toprule
      \textbf{Label} &  \textbf{P} & \textbf{R} & \textbf{F1} & $\kappa$ & \textbf{matches} \\ 
\midrule
          \relLabelPropertyValue &       81.2 &    81.2 & 81.2 & 0.88 &  26 \\ \textit{condSampleFeat.} &       43.1 &    40.0 & 41.5 & 0.23 & 22\\
                 \relLabelUsedIn &       40.0 &    52.2 &  45.3 & 0.55 & 12 \\       \relLabelUsesTechnique &       72.9 &    67.3 & 70.0 & 0.86 & 35\\
                \relLabelHasForm &       54.7 &    71.4 &  61.9 & 0.64  & 35 \\        \relLabelTakenFrom &       33.3 &    63.6 & 43.7 & 0.57 & 7\\
       \textit{measuresProp.} &       80.5 &    72.9 & 76.5 & 0.80 & 70    \\    \relLabelDopedBy &       35.3 &    50.0 & 41.4 & 0.25 & 6\\
      \textit{conditionProp.} &       27.7 &    59.1 & 37.7 &   0.31 & 13  \\      \textit{conditionInstr.} &        24.0 &     60.0 & 34.3 & 0.44 & 6\\
   \textit{conditionEnv.} &        0.0 &     0.0 &    0.0 & 0.00 & 0    \\   \relLabelUsedAs &       23.5 &    85.7 & 36.9 & 0.37 & 12\\
           \relLabelUsedTogether &        13.8 &     28.6 &  18.6 & 0.20 & 4\\ %
\bottomrule
\end{tabular}
    \caption{Inter-annotator agreement: \textbf{relations}.} %
    \label{tab:iaa-relations}
\end{table}

\textbf{Agreement on relations.}
We sample 178 sentences in which each relation occurs at least 25 times according to the first annotator.
We keep NE annotations and ask the second annotator to add relations.
\tref{tab:iaa-relations} shows the results in terms of precision, recall, and $\kappa$ per relation type.
The latter has been computed by treating all pairs of NE annotations as potential relations, using $NO\_REL$ if no relation has been annotated.
Overall $\kappa$ on relations is 0.61 (substantial). %
For each relation label, we can map all other relation types to $OTHER$ and compute agreement for the binary decision whether the label is present or not (analysis suggested by \citet{krippendorff1989content}).
$\kappa$ aims to quantify the degree of agreement \textit{above} chance.
Interpreting our $\kappa$ scores according to the scale of \citet{landis1977application}, we reach at \textbf{fair} agreement for \textit{conditionPropertyValue}, \relLabelUsedTogether, \textit{conditionSampleFeatures}, \relLabelDopedBy, and \relLabelUsedAs.
    We reach \textbf{moderate} agreement for \relLabelUsedIn, \relLabelTakenFrom, and \textit{conditionInstrument}. For the practically important relations \relLabelPropertyValue, \textit{measuresProperty}, and \relLabelUsesTechnique, we even reach \textbf{almost perfect} agreement.

For the non-easily identifiable types, post-hoc discussion with the second annotator (who did not receive an extensive training on the task) concluded it was not always clear to them when using related labels (e.g., \relLabelConditionProperty and \relLabelConditionEnvironment).
Yet, these labels can be learned with good or acceptable accuracy (see \aref{sec:appendix-experimental-results}), indicating that the primary annotator has used the labels consistently.

\section{Task Definitions and Modeling}
\label{sec:models}
In this section, we define several NLP tasks for \corpusNameShort and describe our computational models. %

\subsection{Pre-trained Models}
We use BERT \citep{devlin-etal-2019-bert} as the underlying text encoder for all of our models.
We also use variants of BERT, namely SciBERT \citep{beltagy-etal-2019-scibert}, which has been pre-trained on articles in the scientific domain, and MatSciBERT \citep{gupta_matscibert_2022}, a version of SciBERT further pre-trained on materials science articles.
We use the uncased, 768-dimensional variant of each model, which we fine-tune.

\subsection{Detecting Measurements}
We model the task of classifying whether a sentence contains a \meas or a \qualmeas annotation as a ternary sentence classification task, i.e., it is also possible that a sentence does not refer to any measurement.
As we are primarily interested in detecting \meas, we map the few multi-label cases carrying both positive labels to \meas.
We use a linear layer plus softmax with the \textsc{CLS} token embedding 
as input. %
For training, we downsample the amount of non-measurement sentences. %

\subsection{Named Entity Recognition (NER)}
We compare two state-of-the-art models for NER, (a) a sequence tagger and (b) a dependency parser.
For the \textbf{sequence tagger}, we encode the NE labels using the nested BILOU scheme \citep{alex-etal-2007-recognising}, which leverages a label set of combined types constructed from the training set for nested NEs.
As there are only very few cases (about 0.65\% of all NE annotations) where a token receives more than three stacked NE labels, in order to avoid sparsity issues, we consider only the \enquote{bottom} three layers of stacked entities.
We feed the contextualized embeddings of the last transformer layer of the respective first wordpiece token of each \enquote{real} token into a linear layer and then use a CRF \citep{lafferty2001CRF} to optimize predictions for the entire sequence.

\begin{figure}[t]
    \centering
\includegraphics[width=0.48\textwidth]{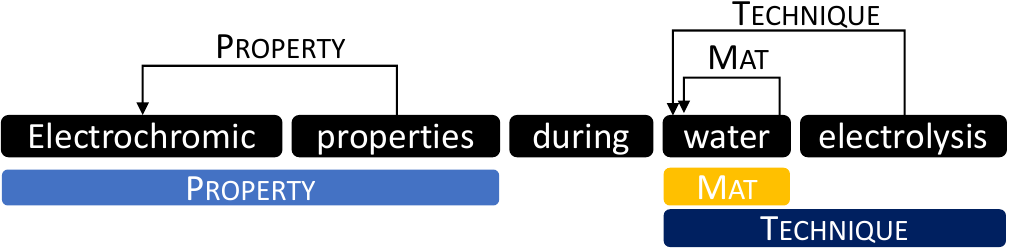}
    \caption{NER as dependency parsing \citep{yu-etal-2020-named}.} %
    \label{fig:dep-parsing-idea}
\end{figure}

Modeling NER as a \textbf{dependency parsing} task \citep{yu-etal-2020-named} can easily account for nested NEs.
The main idea is to predict edges reaching from the end token of an NE to its start token as depicted in \fref{fig:dep-parsing-idea}.
We adapt the STEPS parsing pipeline \citep{grunewald-etal-2021-applying} to the task.
There are three combinations of tags in our dataset that occasionally cover the exact same span and that occur more than 20 times: \textsc{Value}+\neLabelNum, \textsc{Value}+\neLabelRange and \textsc{Mat}+\neLabelForm. %
We hence introduce the above combined labels.
For any other infrequent conflicting labels, we do not add extra tags, i.e., the model can never catch these cases.
We decide on this slight restriction of the model capabilities in order to avoid sparsity issues.
In the evaluation, we do not filter for these cases but of course use all nested NEs as annotated.

\subsection{Relation Extraction}
\label{sec:rel_extraction_model}
Given an input sentence along with all named entities within it, as well as their types (either gold or predicted depending on the experimental setting), we predict which (if any) relation is present between them.
We treat all relations in a single model and predict all relations of a sentence simultaneously by modeling relation extraction as a graph parsing task.
Following \citet{toshniwal-etal-2020-cross}, we first create an embedding $\vec{e}_i$ for the $i$\textsuperscript{th} NE in the sentence by concatenating the token embeddings of its first and last token ($\vec{e}_{i,\textsc{Start}}$, $\vec{e}_{i,\textsc{End}}$).
We also concatenate a learned embedding for the NE's label ($\vec{e}_{i,\textsc{Label}}$):
$\vec{e}_i = \vec{e}_{i,\textsc{Start}} \oplus \vec{e}_{i,\textsc{End}} \oplus \vec{e}_{i,\textsc{Label}}$

Considering NEs as nodes in a graph, we use a biaffine classifier architecture \citep{dozat2017deep} using the implementation of \citet{grunewald-etal-2021-applying,grunewald-etal-2021-robertnlp} to predict the relation between each pair.
The non-existence of a relation is encoded as simply another label ($\varnothing$).
For details on the parser architecture, see \aref{sec:parser-details}.

\section{Experiments}
\label{sec:experiments}
We now detail our experimental results.

\textbf{Experimental Settings.}
We split our corpus into train, dev, and test sets on a per-document basis.
Within the train set, we provide five distinct tune splits (\textit{train1} to \textit{train5}).
For all experiments and for hyperparameter tuning, we always train five models.
Similar to cross-validation, we train on four folds and use the fifth \enquote{training fold} for model selection (cf. \citet{van-der-goot-2021-need} for details). 
Hyperparameters are chosen based on the best dev results, and we finally report results for the test set.
The splits are the same across all tasks.
Because the training data varies across the five runs for which we report results, standard deviations are usually larger than when using the same training data.
For hyperparameter settings, see \aref{sec:appendix-hyperparameters}.

\subsection{Identifying Measurement Sentences}
\tref{tab:meas-identification-small} reports the results for identifying sentences that contain a \neLabelMeasurement or a \textsc{Qual\_Meas} annotation.
In each experiment, we tune the downsampling rate for the majority class \textsc{Other} and the learning rate (using grid search from 1e-4 to 1e-7).
The \textit{random baseline} assigns labels according to the percentage of instances in the (full) training set carrying a particular label.
The average overall accuracy of the MatSciBERT classifier is $78.2\%$.
SciBERT and MatSciBERT perform similarly, with MatSciBERT having a small edge.
Identification of \neLabelMeasurement is comparable to our estimate of human agreement.
For identifying \textsc{Qual\_Meas}, there is headroom.

\begin{table}[t]
    \footnotesize
    \setlength\tabcolsep{1.7pt}
    \begin{tabular}{l|ccc|ccc}
    \toprule
    & \multicolumn{3}{c|}{\neLabelMeasurement} & \multicolumn{3}{c}{\textsc{Qual\_Meas}} \\
    \textbf{LM (sampling)} & \textbf{P} & \textbf{R} & \textbf{F1} & \textbf{P} & \textbf{R} & \textbf{F1}\\
    \midrule
    \textit{Random baseline} & 24.7 & 19.7 & 21.9$_{\pm 2.4}$ & 15.1 & 14.4 & 14.7$_{\pm 1.5}$ \\
         BERT(0.7) & \textbf{74.1} & 71.4 & 72.5$_{\pm2.1}$ & 49.9 & 51.6 & 50.6$_{\pm0.7}$\\
         SciBERT(0.7) & 71.1 & 79.5 & \textbf{75.0}$_{\pm0.7}$ & 52.7 & 52.8 & 52.7$_{\pm1.4}$\\
         MatSciBERT(0.85) & 70.6 & \textbf{80.1} & 74.9$_{\pm1.3}$ &  \textbf{52.8} & \textbf{56.9} & \textbf{54.7}$_{\pm1.0}$\\
         \midrule
         \textit{human agreement*} & \textit{68.6} & \textit{80.9} & \multicolumn{1}{l}{\textit{73.8}} & \textit{61.7} & \textit{61.7} & \multicolumn{1}{l}{\textit{61.7}} \\
         \bottomrule
    \end{tabular}
    \caption{Ternary sentence classification results for \textbf{identifying measurement sentences} on test set. \enquote{Sampling} indicates amount of \textsc{Other} sentences used for training.} \textit{*estimated on subset of data.}
    \label{tab:meas-identification-small}
\end{table}

\begin{table}[t]
    \centering
    \footnotesize
    \setlength\tabcolsep{6pt}
    \begin{tabular}{llrr}
    \toprule
    \textbf{Model} & \textbf{LM} & \textbf{Micro F1} & \textbf{Macro F1} \\
    \midrule
        Dependency & BERT & 73.0$_{\pm 0.7}$ & 61.7$_{\pm 2.3}$\\
         Parser & SciBERT & 76.5$_{\pm 0.3}$ & 65.8$_{\pm 0.9}$\\
         & MatSciBERT & \textbf{77.3$_{\pm 0.3}$} & \textbf{67.6$_{\pm 1.4}$} \\
         \midrule
         CRF Tagger & BERT & 75.3$_{\pm 0.6}$ & 63.2$_{\pm 0.7}$ \\
         {} & SciBERT & 78.7$_{\pm 0.4}$ & 69.3$_{\pm 1.0}$ \\
         {} & MatSciBERT & \textbf{79.6$_{\pm 0.4}$} & \textbf{70.7$_{\pm 0.7}$} \\
    \bottomrule
    \end{tabular}
    \caption{\textbf{Named entity recognition} results on test set.}
    \label{tab:ner_bilou_results-main}
\end{table}

\subsection{Named Entity Recognition Results}
\tref{tab:ner_bilou_results-main} shows the results for named entity recognition.
Again, MatSciBERT performs best with Micro F1 scores approaching 80, which indicates that NE mentions are consistently annotated in \corpusNameShort.
The CRF-based tagger outperforms the dependency-parser-based NER model by a considerable margin. %
For detailed per-label statistics, see \aref{sec:appendix-experimental-results}.
Precision and recall are approximately balanced for all labels.
An exception is \neLabelSample, which is both infrequent in the dataset and hard to identify for humans.
Both models suffer from low recall for this tag.

\begin{table}[t]
    \centering
    \footnotesize
    \setlength\tabcolsep{2pt}
    \begin{tabular}{l|rr|rr}
    \toprule
    & \multicolumn{2}{c|}{\textbf{dev}} & \multicolumn{2}{c}{\textbf{test}}\\
        \textbf{LM} & \textbf{Micro F1} & \textbf{Macro F1} & \textbf{Micro F1} & \textbf{Macro F1} \\
        \midrule
        \textit{Maj. basel.} & 38.3$_{\textcolor{white}{\pm 0.0}}$ & 29.4$_{\textcolor{white}{\pm 0.0}}$ & 37.2$_{\textcolor{white}{\pm 0.0}}$ & 27.4$_{\textcolor{white}{\pm 0.0}}$\\
        BERT & 69.5$_{\pm 0.5 }$ & 63.4$_{\pm 1.0 }$  & 63.5$_{\pm 0.6 }$ & 57.7$_{\pm 1.1 }$ \\
        SciBERT &  72.5$_{\pm 0.8 }$ & 65.7$_{\pm 0.6 }$ & 67.5$_{\pm 0.9 }$ & \textbf{62.0}$_{\pm 2.2 }$ \\
        MatSciBERT & \textbf{73.2}$_{\pm 1.0 }$ & \textbf{66.5}$_{\pm 1.1 }$ & \textbf{67.6}$_{\pm 1.0 }$ & \textbf{62.0}$_{\pm 1.0 }$ \\
        \bottomrule
    \end{tabular}
    \caption{\textbf{Relation extraction} results: gold entities.} %
    \label{tab:rel_extraction_results_test}
\end{table}

\subsection{Relation Extraction Results}
\tref{tab:rel_extraction_results_test} shows the results for relation extraction on gold entities.
A predicted relation is counted as correct if and only if there is a relation with the same start span, end span, and relation label in the set of gold relations for the sentence.
The majority baseline assigns to each pair of entities the relation that is most common in the training set for the respective entity types of the governing and dependent spans (see \aref{sec:appendix-experimental-results}).

The results demonstrate that a biaffine dependency parsing approach achieves robust performance overall and outperforms the baseline by a substantial margin.
The two models trained on scientific text outperform BERT.
Their results are similar, with MatSciBERT having a slight edge.

Analysis of per-label scores (see \aref{sec:appendix-experimental-results}) for MatSciBERT) shows that the highest scores are achieved for \textit{conditionInstrument} (92.2 F1), \textit{usesTechnique} (91.0 F1), and \textit{takenFrom} (84.7).
This is somewhat surprising especially for \textit{conditionInstrument} and \textit{takenFrom}, as these are among the rarest relation types in the corpus (see \tref{tab:relCounts-condensed}).
However, our majority baseline achieves high accuracies on these relation types as well (>90 F1 for \textit{conditionInstrument} and \textit{usesTechnique}), i.e., they are easily inferable from entity types.
The worst performance is observed on the relation types \textit{usedTogether} (4.0 F1), \textit{dopedBy} (22.7 F1), and \textit{usedIn} (37.9 F1). 
These relations occur relatively rarely and also cannot be inferred from entity types. %

\textbf{Relation extraction on predicted entities.}
Finally, we also run our relation extraction module on predicted named entities using the respective best-performing models (both based on MatSciBERT).
Models are evaluated as above, with the additional requirement that the boundaries of start/end spans of a predicted relation must also exactly match those of the respective gold spans.
Prediction accuracy drops substantially: to micro-F1 scores of 42.5 and 36.5 on dev and test, respectively, corresponding to macro-F1 scores of 37.9 and 32.8.
The reason for this is error propagation as relations can only be retrieved if the entities are predicted correctly, and as incorrectly labeled entities can mislead the relation classifier.

\begin{table}[t]
    \centering
    \footnotesize
    \begin{tabular}{llrr}
    \toprule
    \textbf{Dataset} & \textbf{LR} & \textbf{Micro F1} & \textbf{Macro F1} \\
    \midrule
         \corpusNameShort & 1e-4/7e-3 & \textbf{79.6$_{\pm 0.4}$} & \textbf{70.7$_{\pm 0.7}$} \\
         ~~ $+$ SOFC-Exp & {} & 79.5$_{\pm 0.4}$ & 70.4$_{\pm 1.3}$  \\
         ~~ $+$ MSPT & {} & 79.2$_{\pm 0.6}$  & 69.9$_{\pm 0.8}$  \\
         \midrule
         SOFC-Exp & 3e-4/7e-3 & \textbf{83.4}$_{\pm 0.9}$ & \textbf{81.0}$_{\pm 1.0}$ \\
         ~~ $+$ \corpusNameShort & {} & 81.9$_{\pm 1.7}$ & 79.6$_{\pm 1.7}$ \\
         \midrule
         MSPT & 5e-5/9e-3 & \textbf{81.6}$_{\pm 0.4}$ & \textbf{57.8}$_{\pm 0.6}$\\
         ~~ $+$ \corpusNameShort & {} & 80.4$_{\pm 0.4}$ & 56.4$_{\pm 0.8}$ \\
    \bottomrule
    \end{tabular}
    \caption{\textbf{NER MTL} results: MatSciBERT tagger.}
    \label{tab:ner_mt_results}
\end{table}

\subsection{Multi-Task Learning Across Datasets}
\label{sec:mtl-experiments}
To find out whether information extraction accuracy can be increased by employing multi-task learning (MTL), we perform a series of experiments in which we combine \corpusNameShort training data with NE and relation data from other materials science datasets, namely the SOFC-Exp and MSPT corpora (see \sref{sec:relwork}).\footnote{We removed one document from the test split of the SOFC-Exp corpus that is also part of the train set of \corpusNameShort.}
In all experiments, we use a shared MatSciBERT and one classification head for each task (dataset).
When reporting results on \corpusNameShort, we use the same setup as before, but add the complete training sets of SOFC-Exp or MSPT during training.
When reporting results on SOFC and MSPT, we train on all of their training data and the complete training data of \corpusNameShort and perform early stopping on dev.
For these experiments, reported scores are averages over 5 runs with different random seeds.

For \textbf{NER} (\tref{tab:ner_mt_results}), we do not observe overall improvements. %
We hypothesize that this is because in SOFC-Exp, NE types are much coarser-grained, and in MSPT, NE annotations are focused on synthesis procedure paragraphs only.
Nevertheless, as can be seen by the per-label scores in \aref{sec:appendix-experimental-results}, average scores on \corpusNameShort are mainly hurt by decreases on \neLabelSample, while scores for \neLabelRange increase considerably by up to 3.9\%.

\begin{table}
    \centering
    \footnotesize
    \setlength\tabcolsep{3pt}
    \begin{tabular}{l|rr|rr}
    \toprule
    & \multicolumn{2}{c}{\textbf{dev}} & \multicolumn{2}{c}{\textbf{test}}\\
        \textbf{Dataset} & \textbf{Micro F1} & \textbf{Macro F1} & \textbf{Micro F1} & \textbf{Macro F1} \\
        \midrule
        \corpusNameShort & 73.2$_{\pm 1.0 }$ & 66.5$_{\pm 1.1 }$ & 67.6$_{\pm 1.0 }$ & 62.0$_{\pm 1.0 }$ \\
        ~~ + SOFC & 72.5$_{\pm 0.9 }$ & 65.8$_{\pm 0.7 }$ & 68.1$_{\pm 0.7 }$ & 61.1$_{\pm 0.7 }$ \\%\blue{updated} \\
        ~~ + MSPT & \textbf{73.9}$_{\pm 0.4 }$ & \textbf{67.4}$_{\pm 0.4 }$ & \textbf{68.7}$_{\pm 0.7 }$ & \textbf{63.7}$_{\pm 0.7 }$ \\
        \midrule
        SOFC-Exp & 71.3$_{\pm 0.6 }$ & 62.8$_{\pm 2.1 }$ & 66.9$_{\pm 1.6 }$ & 59.8$_{\pm 1.5 }$ \\ %
        ~~ + \corpusNameShort & \textbf{72.3}$_{\pm 0.5 }$ & \textbf{64.3}$_{\pm 3.7 }$ & \textbf{68.7}$_{\pm 1.5 }$ & \textbf{60.9}$_{\pm 3.6 }$ \\ %
        ~~ + MSPT & \textbf{72.3}$_{\pm 0.9 }$ & 63.1$_{\pm 2.6 }$ & 67.6$_{\pm 2.0 }$ & 60.8$_{\pm 4.3 }$ \\ %
        \midrule
        MSPT & 84.2$_{\pm 0.6 }$ & 82.5$_{\pm 0.7 }$ & 84.6$_{\pm 0.8 }$ & 83.0$_{\pm 0.8 }$ \\
        ~~ + \corpusNameShort & \textbf{85.3}$_{\pm 0.2 }$ & \textbf{83.4}$_{\pm 0.8 }$ & \textbf{85.6}$_{\pm 0.4 }$ & \textbf{84.1}$_{\pm 0.6 }$ \\
        ~~ + SOFC & 83.7$_{\pm 1.4 }$ & 81.8$_{\pm 1.4 }$ & 84.7$_{\pm 1.1 }$ & 83.2$_{\pm 1.4 }$ \\ %
        \bottomrule
    \end{tabular}
    \caption{Relation extraction multi-tasking results using MatSciBERT-based parser.}
    \label{tab:rel_extraction_results_mtl}
\end{table}

Results for MTL for \textbf{relations} are shown in \tref{tab:rel_extraction_results_mtl}.
We observe that adding \corpusNameShort to the training data of both SOFC-Exp and MSPT results in improvements.
Incorporating SOFC-Exp instances in the training does not meaningfully increase prediction accuracy on \corpusNameShort, whereas incorporating instances from MSPT leads to modest improvements.
Intuitively, this makes sense: relations in SOFC-Exp focus on a specific type of experiment, while \corpusNameShort covers a broader range of measurements.
Similarly, some \corpusNameShort relations bear resemblance to MSPT relations (e.g., those dealing with instruments or apparatus), which explains why training jointly is beneficial.

\section{Conclusion and Outlook}
\label{sec:conclusion}
In this resource paper, we have presented a new large-scale dataset of 50 scientific articles in the domain of materials science \textit{exhaustively} annotated with named entity mentions, relations, and measurement-related frames.
Our inter-annotator agreement study shows good agreement for most decisions.
Our experiments with state-of-the-art neural models highlight that most distinctions can be learned with good accuracy, and that synergies can be achieved by training jointly with existing more specific materials-science NLP datasets.

Future work is needed to improve on end-to-end or joint models of NER and relation extraction as our experiments showed that a pipeline-based setting suffers from error propagation.
A potential next step is to adapt sequence-to-sequence models to the structure induction tasks of \corpusNameShort, following ideas of \citep{hsu-etal-2022-degree,lu-etal-2021-text2event}.
Finally, employing data augmentation techniques in particular for the less frequent relation types is a viable path for future work.

\section*{Limitations}
As discussed in \sref{sec:iaa-discussion}, we expect our inter-annotator agreement scores to underestimate the reproducibility of the task.
It is, unfortunately, not trivial to find annotators with the required background knowledge.
Hence, scores reflect agreement after only an initial very brief training phase, but nevertheless (in our opinion) give useful insights on the relative difficulty of the labeling decisions.

In our relation extraction experiments, we use label embeddings based on either gold or predicted entity labels (depending on the experimental setup) as an input to our system.
Providing gold entity label information in particular constitutes a setting that is considerably easier for a relation classifier than providing no label information.
Using predicted entity mention and labels showed to suffer from error propagation.
In future work, it may be interesting to evaluate the performance of a relation extraction system that is not given label information, or that predicts entity labels jointly with relations.

\section*{Ethical Considerations}
The annotators participating in our project were completely aware of the goal of the annotations and even helped designing the annotation scheme.
They gave explicit consent to the publication of their annotations.
The main annotator was paid considerably above our country's minimum wage.

\bibliography{custom,anthology}
\bibliographystyle{acl_natbib}

\section*{Appendix}
\appendix

\section{Hyperparameters}
\label{sec:appendix-hyperparameters}

We use AdamW \citep{loshchilov2019decoupled} as optimizer for all our models.
We use an inverse square-root learning rate scheduler similar to the one used by \citet{vaswani2017} where $ws$ refers to the number of warmup steps:

\begin{center}
{\small
    $lr = \sqrt{ws} \cdot \min(\frac{1}{\sqrt{step\_num}}, step\_num \cdot ws^{-1.5})$
    }
\end{center}

For our measurement identification experiments, we downsample the amount of non-measurement sentences since they represent the majority in the training data.
We tune this downsampling rate per model since each BERT variant has been shown to prefer a slightly different one.
We apply early stopping after 3 epochs without improvement in terms of F1.

Our NER sequence tagging models are trained with two separate learning rates; one for BERT + linear output layer and another one for the CRF output layer.
Both learning rates are reported in the respective column of \tref{tab:ner_bilou_hyper}.
We train for 60-100 epochs, depending on the size of the combined dataset, and take the model with the best evaluation score during this period.

\begin{table}[!h]
    \centering
    \footnotesize
    \begin{tabular}{llr}
    \toprule
    \textbf{Model} & \textbf{LM} & \textbf{LR} \\
    \midrule
         CRF & BERT & 1e-4/7e-3 \\
         {} & SciBERT & 5e-5/7e-3  \\
         {} & MatSciBERT & 1e-4/7e-3 \\
         \midrule
         Dep. Pars. & BERT & 2e-4 \\
         {} & SciBERT & 9e-5 \\
         {} & MatSciBERT & 3e-4 \\
    \bottomrule
    \end{tabular}
    \caption{\textbf{Hyperparameters:} Learning rates for NER models.}
    \label{tab:ner_bilou_hyper}
\end{table}

For relation extraction, we use a base learning rate of 4e-5 for all experiments, which we found to perform best in preliminary experiments.
We employ early stopping with a patience of 15 epochs for all experiments.

Our models are trained with Nvidia A100 and V100 GPUs using the PyTorch framework.

\section{Details on Biaffine Parser Architecture}
\label{sec:parser-details}
We here describe the biaffine parser architecture used to predict relations between named entities.
Taking as input the NE embeddings described in \sref{sec:rel_extraction_model}, head and dependent representations for the $i$'th NE are computed via two single-layer feedforward neural networks:
\begin{align*}
	\vec{h}^{head}_i &= \text{FNN}^{head}(\vec{e}_i)\\
	\vec{h}^{dep}_i &= \text{FNN}^{dep}(\vec{e}_i)
\end{align*}

These representations are then fed to a biaffine classifier that maps head--dependent pairs onto logit vectors $s_{i,j}$ whose dimensionality corresponds to the inventory of relation labels.
Using the softmax operation, these scores are transformed into probability distributions $P(y_{i,j})$ over relation labels:
\begin{align*}
    \hspace*{-1mm}\text{Biaff}(\vec{x}_1, \vec{x}_2) &= \vec{x}^\top_1 \vec{U} \vec{x}_2 + W(\vec{x}_1 \oplus \vec{x}_2) + \vec{b} \label{eq:biaff}\\
    \vec{s}_{i,j} &= \text{Biaff}\big( \vec{h}^{head}_i, \vec{h}^{dep}_j \big)\\
	P(y_{i,j}) &= \text{softmax}(\vec{s}_{i,j})
\end{align*}
The predicted relation for a pair of named entities is the one receiving the highest probability (which may be $\varnothing$, i.e., no relation).

\paragraph{Token embeddings.}
The token embeddings $\vec{e}_{i,\textsc{Start}}$ and $\vec{e}_{i,\textsc{End}}$, which form part of the NE embeddings $\vec{e}_i$, are computed as a learned scalar mixture of BERT layers as described by \citet{kondratyuk-straka-2019-75}.

\section{Detailed Corpus Statistics}
\label{sec:app-detailed-corpus-stats}

\tref{tab:ner-counts} shows the NE counts in \corpusNameShort by datasplit.

\tref{tab:iaa-named-entities-appendix} and \tref{tab:iaa-relations-appendix} show the detailed counts for our inter-annotator agreement study.

\textbf{Choice of agreement metrics for evaluating agreement on named entity annotations.}
The task of identifying and labeling NE mentions is a sequence labeling task, hence, $\kappa$ is not applicable. \citet{brandsen-etal-2020-creating} provide a good explanation of why this is the case in their section 5.1.
Using unitizing $\alpha_U$ is an option, but there is no standard implementation or interpretation for NE annotations in the NLP community, and it does not work for overlapping annotations (which we have in our dataset). We opted for using precision and recall, which are intuitively interpretable (\textit{How many of the instances of one type marked by one annotator have also been marked by the respective other annotator?}).
\citet{hripcsak2005agreement} convincingly argue (with a very simple proof) that for sequence labeling tasks such as NEs, F1 actually approaches $\kappa$.

\begin{table}[h!]
\footnotesize
\centering
\begin{tabular}{lrrrr}
\toprule
\textbf{Label} &   \textbf{total}  &  \textbf{train}  &  \textbf{dev}  &  \textbf{test} \\
\midrule
\textsc{Mat} &   15596  &   10875  &  2318  &   2403 \\
\textsc{Num} &    6081  &    4142  &  1077  &    862 \\
\textsc{Value} &    4852  &    3266  &   895  &    691 \\
\textsc{Unit} &    4330  &    2880  &   789  &    661 \\
\textsc{Property} &    3925  &    2867  &   598  &    460 \\
\textsc{Form} &    3568  &    2716  &   345  &    507 \\
\textsc{Measurement} &    2171  &    1531  &   345  &    295 \\
\textsc{Cite} &    1709  &    1280  &   274  &    155 \\
\textsc{Sample} &    1461  &    1031  &   249  &    181 \\
\textsc{Technique} &    1036  &     755  &   146  &    135 \\
\textsc{Dev} &     808  &     459  &   235  &    114 \\
\textsc{Range} &     736  &     546  &   105  &     85 \\
\textsc{Instrument} &     378  &     278  &    50  &     50 \\
\bottomrule
\end{tabular}
\caption{Label counts for \textbf{named entities} in \corpusNameShort.}
\label{tab:ner-counts}
\end{table}

\begin{table}[t]
    \centering
    \footnotesize
    \setlength\tabcolsep{2pt}
\begin{tabular}{lrrrrrr}
\toprule
      \textbf{Label} &  \textbf{P} &  \textbf{R} &  \textbf{matches} 
    & \textbf{matches} &  \# \textbf{A1} &  \# \textbf{A2} \\
    & & & \textbf{exact} & \textbf{relaxed} & \\
\midrule
       MAT &       96.7 &    91.2 &            144 &              145 &           150 &           159 \\
        NUM &       98.9 &   100.0 &             86 &               86 &            87 &            86 \\
      VALUE &      100.0 &   100.0 &             53 &               54 &            54 &            54 \\
       UNIT &       97.9 &   100.0 &             47 &               47 &            48 &            47 \\
   PROPERTY &       42.6 &    37.7 &             21 &               29 &            68 &            77 \\
       FORM &       95.7 &    86.3 &             44 &               44 &            46 &            51 \\
\textsc{Meas.} &       44.6 &    51.0 &             17 &               25 &            56 &            49 \\
       CITE &       97.4 &    97.4 &             38 &               38 &            39 &            39 \\
     SAMPLE &        3.1 &    12.5 &              1 &                1 &            32 &             8 \\
  TECHNIQUE &       77.5 &    59.6 &             25 &               31 &            40 &            52 \\
        DEV &       96.0 &    82.8 &             18 &               24 &            25 &            29 \\
      RANGE &      100.0 &   100.0 &             24 &               25 &            25 &            25 \\
 INSTRUMENT &       80.0 &    76.9 &             18 &               20 &            25 &            26 \\
\bottomrule
\end{tabular}
    \caption{Inter-annotator agreement: \textbf{named entities}. Precision and recall computed from relaxed matches.}
    \label{tab:iaa-named-entities-appendix}
\end{table}

\begin{table}[t]
    \centering
    \footnotesize
    \setlength\tabcolsep{2pt}
\begin{tabular}{lrr|lrr}
\toprule
      \textbf{Label} &  \textbf{P} &  \textbf{R} &  \textbf{matches} &  \# \textbf{A1} &  \# \textbf{A2} \\
\midrule
         propertyValue &       81.2 &    81.2 &             26 &            32 &            32 \\
                 usedIn &       40.0 &    52.2 &             12 &            30 &            23 \\
                hasForm &       54.7 &    71.4 &             35 &            64 &            49 \\
       measuresProperty &       80.5 &    72.9 &             70 &            87 &            96 \\
      conditionProperty &       27.7 &    59.1 &             13 &            47 &            22 \\
   conditionEnvironment &        0.0 &     0.0 &              0 &            19 &             4 \\
           usedTogether &       13.8 &    28.6 &              4 &            29 &            14 \\
conditionSampleFeatures &       43.1 &    40.0 &             22 &            51 &            55 \\
          usesTechnique &       72.9 &    67.3 &             35 &            48 &            52 \\
              takenFrom &       33.3 &    63.6 &              7 &            21 &            11 \\
                dopedBy &       35.3 &    50.0 &              6 &            17 &            12 \\
    conditionInstrument &       24.0 &    60.0 &              6 &            25 &            10 \\
                 usedAs &       23.5 &    85.7 &             12 &            51 &            14 \\
\bottomrule
\end{tabular}
    \caption{Inter-annotator agreement for \textbf{relations}.}
    \label{tab:iaa-relations-appendix}
\end{table}

\section{SOFC-Exp and MSPT Corpora}
In \sref{sec:mtl-experiments}, we perform several multi-task learning (MTL) experiments with \corpusNameShort and two additional NLP datasets in the materials science domain, SOFC-Exp \citep{friedrich-etal-2020-sofc} and MSTP \citep{mysore-etal-2019-materials}.
We here describe them briefly.

There are 4 named entities in the SOFC-Exp corpus: \sofcNeLabelMaterial, which refers to mentions of materials or chemical formulas, \sofcNeLabelValue, which denotes numerical values and their corresponding physical unit, \sofcNeLabelDevice, which marks device types used in an experiment, and \sofcNeLabelExperiment, which indicates frame evoking words.
Furthermore, there are 16 distinct slots that are modeled as relations between experiment frame evoking word and corresponding entity. %
\tref{tab:sofc_relation_counts} shows the counts for these relations.
These counts are not equal to the ones reported by \citet{friedrich-etal-2020-sofc} since we have to remove relations that span across multiple sentences as this case cannot be handled by our relation extraction pipeline.

\begin{table}[htb]
    \centering
    \footnotesize
    \begin{tabular}{lccc}
    \toprule
        \textbf{Label} &  \textbf{train} & \textbf{dev} & \textbf{test}\\
        \midrule
        \sofcRelLabelAnodeMaterial & 220 & 32 & 26 \\
        \sofcRelLabelCathodeMaterial & 173 & 71 & 37 \\
        \sofcRelLabelConductivity & 36 & 19 & 23 \\
        \sofcRelLabelCurrentDensity & 59 & 6 & 17\\
        \sofcRelLabelDegradationRate & 15 & 4 & 1\\
        \sofcRelLabelDevice & 311 & 59 & 109 \\
        \sofcRelLabelElectrolyteMaterial & 187 & 22 & 120\\
        \sofcRelLabelFuelUsed & 124 & 28 & 40\\
        \sofcRelLabelInterlayerMaterial & 34 & 17 & 6\\
        \sofcRelLabelOpenCircuitVoltage & 41 & 3 & 25\\
        \sofcRelLabelPowerDensity & 138 & 24 & 70\\
        \sofcRelLabelResistance & 118 & 15 & 57\\
        \sofcRelLabelSupportMaterial & 88 & 13 & 2\\
        \sofcRelLabelTimeOfOperation & 42 & 3 & 12\\
        \sofcRelLabelVoltage & 30 & 3 & 14\\
        \sofcRelLabelWorkingTemperature & 330 & 63 & 138\\
        \bottomrule
    \end{tabular}
    \caption{SOFC-Exp relation counts in our setup.}
    \label{tab:sofc_relation_counts}
\end{table}

The MSPT corpus introduces additional the 21 entities:
\begin{itemize}
\setlength\itemsep{0pt}
    \item \msptNeLabelPropertyMisc 
    \item \msptNeLabelPropertyUnit
    \item \msptNeLabelNumber 
    \item \msptNeLabelCharacterizationApparatus 
    \item \msptNeLabelApparatusUnit 
    \item \msptNeLabelConditionMisc 
    \item \msptNeLabelMeta 
    \item \msptNeLabelSynthesisApparatus 
    \item \msptNeLabelOperation 
    \item \msptNeLabelAmountMisc 
    \item \msptNeLabelAmountUnit 
    \item \msptNeLabelReference 
    \item \msptNeLabelPropertyType 
    \item \msptNeLabelMaterial 
    \item \msptNeLabelMaterialDescriptor 
    \item \msptNeLabelApparatusDescriptor 
    \item \msptNeLabelApparatusPropertyType 
    \item \msptNeLabelConditionUnit 
    \item \msptNeLabelNonrecipeMaterial 
    \item \msptNeLabelConditionType 
    \item \msptNeLabelBrand
\end{itemize}

\tref{tab:mspt_relation_counts} lists the counts of the 14 relations of the MSPT dataset that we use in our MTL experiments.

\begin{table}[h!]
    \centering
    \footnotesize
    \begin{tabular}{lccc}
    \toprule
        \textbf{Label} &  \textbf{train} & \textbf{dev} & \textbf{test}\\
        \midrule
        \msptRelLabelRecipeTarget & 270 & 53 & 92\\
        \msptRelLabelSolventMaterial & 352 & 61 & 107\\
        \msptRelLabelAtmosphericMaterial & 144 & 25 & 35\\
        \msptRelLabelRecipePrecursor & 654 & 152 & 199\\
        \msptRelLabelParticipantMaterial & 1315 & 236 & 400\\
        \msptRelLabelApparatusOf & 358 & 56 & 93\\
        \msptRelLabelConditionOf & 1378 & 232 & 415\\
        \msptRelLabelDescriptorOf & 1157 & 193 & 333\\
        \msptRelLabelNumberOf & 2114 & 422 & 663\\
        \msptRelLabelAmountOf & 1099 & 244 & 376\\
        \msptRelLabelApparatusAttrOf & 66 & 56 & 24\\
        \msptRelLabelBrandOf & 326 & 83 & 91\\
        \msptRelLabelCoreOf & 177 & 23 & 89\\
        \msptRelLabelNextOperation & 1311 & 233 & 391\\
        \bottomrule
    \end{tabular}
    \caption{MSPT relation counts in our setup.} %
    \label{tab:mspt_relation_counts}
\end{table}

\section{Detailed Experimental Results}
\label{sec:appendix-experimental-results}

This appendix provides further details on our experimental results.
\tref{tab:meas_class_results} depicts the results for identifying sentences containing \neLabelMeasurement or \textsc{Qual\_Meas} annotations.

\textbf{NER.}
\tref{tab:ne_results_1} and \tref{tab:ner-class-statistics} report F1 for NER on \corpusNameShort per label.
\tref{tab:mulms-multi-task-per-label-scores} gives per-label scores for NER in our MTL experiments.
\tref{tab:sofc-multi-task-per-label-scores} and \tref{tab:mspt-multi-task-per-label-scores} provide per-label scores for the SOFC-Exp corpus and MSPT corpus in a single-task setting as well as in a multi-task setting with \corpusNameShort added to the training.

\textbf{Relation extraction.}
\tref{tab:all_relation_results_test} lists per-relation scores when using gold NEs or when using predicted NEs for relation extraction, as well as per-relation scores for the majority baseline.
\tref{tab:rel_extraction_results_perlabel} shows relation extraction scores per label for both dev and test.
\tref{tab:ner_rel_pipeline_results} shows overall results for predicted entities on dev and test.

\begin{table}[htb]
    \centering
    \footnotesize
\begin{tabular}{lccc}
\toprule
\textbf{Label} &       \textbf{P} &       \textbf{R} &      \textbf{F1} \\
\midrule
\neLabelMat         &  83.6 &  82.2 &  82.8 \\[-0.15cm]
         & \tiny{ $ \pm 1.3 $ } & \tiny{ $ \pm 1.2 $ } & \tiny{ $ \pm 0.8 $ } \\
\neLabelNum         &  94.9 &  94.8 &  94.9 \\[-0.15cm]
         & \tiny{ $ \pm 0.6 $ } & \tiny{ $ \pm 1.0 $ } & \tiny{ $ \pm 0.7 $ } \\
\neLabelValue       &  89.4 &  87.0 &  88.2 \\[-0.15cm]
         & \tiny{ $ \pm 0.6 $ } & \tiny{ $ \pm 1.2 $ } & \tiny{ $ \pm 0.9 $ } \\
\neLabelUnit        &  94.2 &  90.4 &  92.3 \\[-0.15cm]
         & \tiny{ $ \pm 0.4 $ } & \tiny{ $ \pm 1.1 $ } & \tiny{ $ \pm 0.6 $ } \\
\neLabelProperty    &  49.8 &  53.0 &  51.1 \\[-0.15cm]
         & \tiny{ $ \pm 3.0 $ } & \tiny{ $ \pm 4.5 $ } & \tiny{ $ \pm 1.4 $ } \\
\neLabelCite        &  88.6 &  87.7 &  88.2 \\[-0.15cm]
         & \tiny{ $ \pm 0.8 $ } & \tiny{ $ \pm 2.0 $ } & \tiny{ $ \pm 1.3 $ } \\
\neLabelTechnique   &  49.6 &  51.1 &  50.1 \\[-0.15cm]
         & \tiny{ $ \pm 3.4 $ } & \tiny{ $ \pm 5.6 $ } & \tiny{ $ \pm 2.9 $ } \\
\neLabelRange       &  70.3 &  74.8 &  72.3 \\[-0.15cm]
         & \tiny{ $ \pm 5.8 $ } & \tiny{ $ \pm 3.5 $ } & \tiny{ $ \pm 3.0 $ } \\
\neLabelInstrument  &  46.7 &  44.8 &  45.6 \\[-0.15cm]
         & \tiny{ $ \pm 2.7 $ } & \tiny{ $ \pm 3.5 $ } & \tiny{ $ \pm 2.0 $ } \\
\neLabelSample      &  72.5 &  36.7 &  47.9 \\[-0.15cm]
         & \tiny{ $ \pm 10.2 $ } & \tiny{ $ \pm 6.2 $ } & \tiny{ $ \pm 4.1 $ } \\
\neLabelForm        &  66.5 &  71.4 &  68.9 \\[-0.15cm]
         & \tiny{ $ \pm 3.0 $ } & \tiny{ $ \pm 1.9 $ } & \tiny{ $ \pm 2.5 $ } \\
\neLabelDev         &  82.6 &  74.9 &  78.6 \\[-0.15cm]
         & \tiny{ $ \pm 2.0 $ } & \tiny{ $ \pm 2.5 $ } & \tiny{ $ \pm 1.9 $ } \\
\neLabelMeasurement &  61.6 &  55.3 &  58.2 \\[-0.15cm]
         & \tiny{ $ \pm 2.1 $ } & \tiny{ $ \pm 3.0 $ } & \tiny{ $ \pm 0.8 $ } \\
\bottomrule
\end{tabular}
\caption{Per label scores for NER using BILOU tagging and MatSciBERT.}
\label{tab:ne_results_1}
\end{table}

\begin{table}[htb]
    \centering
    \footnotesize
    \begin{tabular}{lcc}
    \toprule
        \textbf{Label} &  \textbf{CRF-Tagger} & \textbf{Dep. Parser}\\
        \midrule
        \neLabelMat & \textbf{82.8}$_{\pm0.8}$  & 80.0$_{\pm0.4}$\\
        \neLabelNum & \textbf{94.9}$_{\pm0.7}$  & 94.2$_{\pm0.4}$\\
        \neLabelValue & \textbf{88.2}$_{\pm0.9}$  & 82.7$_{\pm1.1}$\\
        \neLabelUnit & \textbf{92.3}$_{\pm0.6}$  & 90.8$_{\pm0.7}$\\
        \neLabelProperty & 51.1$_{\pm1.4}$  & \textbf{51.7}$_{\pm2.0}$\\
        \neLabelCite & \textbf{88.2}$_{\pm1.3}$  & 85.7$_{\pm1.5}$\\
        \neLabelTechnique & 50.1$_{\pm2.9}$  & \textbf{51.4}$_{\pm2.4}$\\
        \neLabelRange & \textbf{72.3}$_{\pm 3.0}$ & 66.4$_{\pm 4.0}$\\
        \neLabelInstrument & \textbf{45.6}$_{\pm2.0}$  & 44.1$_{\pm3.3}$\\
        \neLabelSample & \textbf{47.9}$_{\pm4.1}$  & \hspace*{1mm}29.4$_{\pm18.1}$\\
        \neLabelForm & \textbf{68.9}$_{\pm2.5}$  & 67.6$_{\pm1.1}$\\
        \neLabelDev & \textbf{78.6}$_{\pm1.9}$  & 76.4$_{\pm2.2}$\\
        \neLabelMeasurement & 58.2$_{\pm0.8}$  & \textbf{58.7}$_{\pm0.7}$\\
        \bottomrule
    \end{tabular}
    \caption{\textbf{Per-Label NER} results on test in terms of \textbf{F1} (using MatSciBERT).} %
    \label{tab:ner-class-statistics}
\end{table}

\begin{table}[htb]
    \centering
    \footnotesize
    \begin{tabular}{lccc}
    \toprule
        \textbf{Label} &  \textbf{Single-Task} & \textbf{+ SOFC} & \textbf{+ MSPT}\\
        \midrule
        \neLabelMat & \textbf{82.8} & 82.3 & 82.3 \\
        \neLabelNum & 94.9 & \textbf{95.6} & \textbf{95.6} \\
        \neLabelValue & \textbf{88.2} &  88.1 & \textbf{88.2}\\
        \neLabelUnit & 92.3  & 92.1 & \textbf{92.6}\\
        \neLabelProperty & \textbf{51.1} & 49.4 & 50.8\\
        \neLabelCite & \textbf{88.2} & 88.1 & 87.8\\
        \neLabelTechnique & 50.1 & \textbf{52.9} & 50.0\\
        \neLabelRange & 72.3 & \textbf{76.2} & 75.2\\
        \neLabelInstrument & 45.6 & \textbf{45.9} & 42.8\\
        \neLabelSample & \textbf{47.9} & 38.3 & 45.8\\
        \neLabelForm & 68.9 & \textbf{69.9} & 64.9\\
        \neLabelDev & \textbf{78.6} & 77.9 & 76.3\\
        \neLabelMeasurement & \textbf{58.2}  & 57.8 & 56.9\\
        \bottomrule
    \end{tabular}
    \caption{\textbf{Per-Label NER} results for \corpusNameShort on test in terms of \textbf{F1} for single-task and multi-task MatSciBERT taggers.}
    \label{tab:mulms-multi-task-per-label-scores}
\end{table}

\begin{table}[htb]
    \centering
    \footnotesize
    \begin{tabular}{lcc}
    \toprule
        \textbf{Label} &  \textbf{Single-Task} & \textbf{+ \corpusNameShort}\\
        \midrule
        \sofcNeLabelMaterial & \textbf{75.8} & 73.2 \\
        \sofcNeLabelExperiment & \textbf{81.7} & 81.2\\
        \sofcNeLabelValue & \textbf{93.9} & 92.0 \\
        \sofcNeLabelDevice & \textbf{72.6} & 72.0 \\
        \bottomrule
    \end{tabular}
    \caption{\textbf{Per-Label Named Entity Recognition} results for SOFC-Exp on test in terms of \textbf{F1} using single-task and multi-task MatSciBERT taggers.}
    \label{tab:sofc-multi-task-per-label-scores}
\end{table}

\begin{table}[htb]
    \centering
    \footnotesize
    \setlength\tabcolsep{3pt}
    \begin{tabular}{lcc}
    \toprule
        \textbf{Label} &  \textbf{ST} & \textbf{+ \corpusNameShort}\\
        \midrule
        \msptNeLabelMeta                      & \textbf{47.5} & 46.3 \\
        \msptNeLabelPropertyMisc              & 32.8 & \textbf{34.7} \\
        \msptNeLabelSynthesisApparatus        & \textbf{68.7} & 66.7 \\
        \msptNeLabelOperation                 & \textbf{85.0} & 84.9 \\
        \msptNeLabelPropertyUnit              & 42.3 & \textbf{44.5} \\
        \msptNeLabelAmountMisc                & \textbf{41.4} & 26.0 \\
        \msptNeLabelNumber                    & 94.8 & \textbf{95.5} \\
        \msptNeLabelAmountUnit                & \textbf{95.5} & 95.0 \\
        \msptNeLabelReference                 & \textbf{70.9} & 67.7 \\
        \msptNeLabelPropertyType              & \textbf{24.6} & 19.0 \\
        \msptNeLabelMaterial                  & \textbf{84.1} & 81.6 \\
        \msptNeLabelMaterialDescriptor        & \textbf{67.8} & 63.7 \\
        \msptNeLabelCharacterizationApparatus & 16.2 & \textbf{28.8} \\
        \msptNeLabelApparatusUnit             & 57.8 & \textbf{61.5} \\
        \msptNeLabelApparatusDescriptor       & \textbf{67.0} & 65.1 \\
        \msptNeLabelApparatusPropertyType     & 0.0  & 0.0  \\
        \msptNeLabelConditionMisc             & 72.3 & \textbf{73.5} \\
        \msptNeLabelConditionUnit             & \textbf{95.2} & 94.3 \\
        \msptNeLabelNonrecipeMaterial         & \textbf{62.3} & 59.6 \\
        \msptNeLabelConditionType             & \textbf{15.7} & 12.8 \\
        \msptNeLabelBrand                     & \textbf{71.1} & 64.0 \\
        \bottomrule
    \end{tabular}
    \caption{\textbf{Per-Label Named Entity Recognition} results for MSPT on test in terms of \textbf{F1} using single-task (ST) and multi-task MatSciBERT taggers.}
    \label{tab:mspt-multi-task-per-label-scores}
\end{table}

\begin{table}[htb]
    \centering
    \footnotesize
    \setlength\tabcolsep{3pt}
    \begin{tabular}{lcc}
    \toprule
    & \textbf{micro F1} & \textbf{macro F1}\\
    \midrule
    \textbf{dev} & 42.5$_{\pm 1.0 }$ & 37.9$_{\pm 1.7 }$ \\
    \textbf{test} & 36.5$_{\pm 0.9 }$ & 32.8$_{\pm 1.2 }$ \\
    \bottomrule
    \end{tabular}
    \caption{Relation extraction results in terms of \textbf{F1}, predicted named entities (including standard deviation over five folds).}
    \label{tab:ner_rel_pipeline_results}
\end{table}

\begin{table}[htb]
    \centering
    \footnotesize
    \begin{tabular}{lcc}
    \toprule
    \textbf{Label} & \textbf{dev} & \textbf{test} \\
    \midrule
    \textit{hasForm} & 71.3$_{\pm 0.8}$ & 76.1$_{\pm 0.5}$ \\
    \textit{measuresProperty} & 88.0$_{\pm 1.0}$ & 83.1$_{\pm 0.8}$ \\
    \textit{measuresPropertyValue} & 84.1$_{\pm 2.2}$ & 73.8$_{\pm 1.0}$ \\
    \textit{usedAs} & 50.2$_{\pm 2.3}$ & 41.8$_{\pm 1.9}$ \\
    \textit{conditionProperty} & 83.0$_{\pm 1.3}$ & 72.3$_{\pm 1.0}$ \\
    \textit{conditionPropertyValue} & 74.7$_{\pm 2.5}$ & 63.2$_{\pm 2.5}$ \\
    \textit{conditionSampleFeatures} & 67.6$_{\pm 0.9}$ & 66.0$_{\pm 2.1}$ \\
    \textit{usesTechnique} & 94.6$_{\pm 0.6}$ & 91.0$_{\pm 0.9}$ \\
    \textit{conditionEnvironment} & 57.1$_{\pm 6.2}$ & 39.0$_{\pm 3.0}$ \\
    \textit{propertyValue} & 86.7$_{\pm 1.3}$ & 82.5$_{\pm 2.1}$ \\
    \textit{usedIn} & 49.3$_{\pm 5.6}$ & 37.9$_{\pm 3.7}$ \\
    \textit{conditionInstrument} & 98.2$_{\pm 0.8}$ & 92.2$_{\pm 0.9}$ \\
    \textit{dopedBy} & \textcolor{white}{0}0.0$_{\pm 0.0}$ & 22.7$_{\pm 18.7}$ \\
    \textit{takenFrom} & 85.5$_{\pm 3.7}$ & 84.7$_{\pm 3.6}$ \\
    \textit{usedTogether} & \textcolor{white}{0}7.5$_{\pm 2.0}$ & \textcolor{white}{0}4.0$_{\pm 1.7}$ \\
    \midrule
    Macro-avg. & 66.5$_{\pm 1.1}$ & 62.0$_{\pm 1.0}$\\
    Micro-avg. & 73.2$_{\pm 1.0}$ & 67.6$_{\pm 1.0}$\\
    \bottomrule
    \end{tabular}
    \caption{Per-label F1 scores for relation extraction using MatSciBERT (gold named entities).}
    \label{tab:rel_extraction_results_perlabel}
\end{table}

\begin{table*}[htb]
    \centering
    \footnotesize
    \setlength\tabcolsep{7pt}
    \begin{tabular}{ll|ccc|ccc}
        \toprule
        {} & {} & \multicolumn{3}{c}{\textbf{dev}} & \multicolumn{3}{c}{\textbf{test}}\\
         \textbf{LM} & \textbf{Label} & \textbf{P} & \textbf{R} & \textbf{F1} & \textbf{P} & \textbf{R} & \textbf{F1} \\
         \midrule
         BERT & 
         \threerows{\meas}{\qualmeas}{2-Class Macro Avg.} & 
         \threerows{\textbf{71.5}$_{\pm 1.9}$}{$45.6_{\pm 2.8}$}{$58.5_{\pm 2.0}$} & 
         \threerows{$67.1_{\pm 3.0}$}{$61.4_{\pm 3.7}$}{$64.3_{\pm 2.8}$} & 
         \threerows{$69.2_{\pm 1.3}$}{$52.2_{\pm 2.1}$}{$60.7_{\pm 1.3}$} &
         
         \threerows{\textbf{74.1}$_{\pm 3.1}$}{$49.9_{\pm 3.1}$}{\textbf{62.0}$_{\pm 3.0}$} &
         \threerows{$71.4_{\pm 5.6}$}{$51.6_{\pm 3.2}$}{$61.5_{\pm 3.7}$} &
         \threerows{$72.5_{\pm 2.1}$}{$50.6_{\pm 0.7}$}{$61.5_{\pm 1.2}$}\\
         \midrule
         SciBERT & 
         \threerows{\meas}{\qualmeas}{2-Class Macro Avg.} & 
         \threerows{69.1$_{\pm 1.9}$}{\textbf{54.0}$_{\pm 1.7}$}{\textbf{61.5}$_{\pm 1.2}$} & 
         \threerows{$77.6_{\pm 1.5}$}{66.3$_{\pm 5.0}$}{72.0$_{\pm 2.8}$} & 
         \threerows{$73.1_{\pm 0.4}$}{\textbf{59.4}$_{\pm 1.9}$}{\textbf{66.2}$_{\pm 1.0}$} &
         
         \threerows{$71.1_{\pm 2.0}$}{$52.7_{\pm 0.5}$}{$61.9_{\pm 0.9}$} &
         \threerows{$79.5_{\pm 1.6}$}{52.8$_{\pm 3.2}$}{$66.2_{\pm 1.8}$} &
         \threerows{\textbf{75.0}$_{\pm 0.7}$}{$52.7_{\pm 1.4}$}{$63.9_{\pm 0.6}$}\\
         \midrule
         MatSciBERT & 
         \threerows{\meas}{\qualmeas}{2-Class Macro Avg.} &
         \threerows{$69.4_{\pm 2.6}$}{51.4$_{\pm 1.4}$}{60.4$_{\pm 1.9}$} & 
         \threerows{\textbf{77.9}$_{\pm 4.0}$}{\textbf{67.0}$_{\pm 1.6}$}{\textbf{72.4}$_{\pm 2.0}$} & 
         \threerows{\textbf{73.2}$_{\pm 0.7}$}{$58.2_{\pm 1.1}$}{$65.7_{\pm 0.4}$} &
         
         \threerows{70.6$_{\pm 2.2}$}{\textbf{52.8}$_{\pm 0.8}$}{61.7$_{\pm 1.0}$} &
         \threerows{\textbf{80.1}$_{\pm 3.5}$}{\textbf{56.9}$_{\pm 2.6}$}{\textbf{68.5}$_{\pm 2.4}$} &
         \threerows{74.9$_{\pm 0.6}$}{\textbf{54.7}$_{\pm 1.0}$}{\textbf{64.8}$_{\pm 0.7}$} \\
         \midrule
         \textit{human} & \meas & & & \textit{74.2} & & & \textit{74.2} \\
         \textit{agreement} & \qualmeas & & & \textit{61.7} & & & \textit{61.7}\\
         \bottomrule
    \end{tabular}
    \caption{Ternary sentence classification results for identifying sentences containing \meas or \qualmeas annotations vs. \None.
    Human agreement is only suitable for a rough comparison because it is estimated on a subset of the data.} %
    \label{tab:meas_class_results}
\end{table*}

\begin{table*}[htb]
    \centering
    \footnotesize
    \begin{tabular}{lccc|ccc|ccc}
    \toprule
     & \multicolumn{3}{c}{\textbf{Gold Entities}} & \multicolumn{3}{c}{\textbf{Predicted Entities}} &  \multicolumn{3}{c}{\textbf{Majority Baseline}}\\
    \textbf{Label} & \textbf{P} & \textbf{R} & \textbf{F1} & \textbf{P} & \textbf{R} & \textbf{F1} & \textbf{P} & \textbf{R} & \textbf{F1} \\
     \midrule
    \textit{hasForm} & 74.0 & 78.5 & 76.1 & 54.2 & 55.5 & 54.8 & 0.0 & 0.0 & 0.0 \\[-0.15cm]
    & \tiny{ $ \pm 1.8 $ } & \tiny{ $ \pm 1.1 $ } & \tiny{ $ \pm 0.5 $ } & \tiny{ $ \pm 3.6 $ } & \tiny{ $ \pm 1.6 $ } & \tiny{ $ \pm 2.3 $ } \\
    \textit{measuresProperty} & 80.2 & 86.3 & 83.1 & 39.3 & 36.6 & 37.8 & 50.5 & 99.6 & 67.0 \\[-0.15cm]
    & \tiny{ $ \pm 1.1 $ } & \tiny{ $ \pm 2.2 $ } & \tiny{ $ \pm 0.8 $ } & \tiny{ $ \pm 2.8 $ } & \tiny{ $ \pm 2.2 $ } & \tiny{ $ \pm 1.4 $ } \\
    \textit{measuresPropertyValue} & 68.6 & 80.0 & 73.8 & 41.5 & 38.3 & 39.7 & 0.0 & 0.0 & 0.0 \\[-0.15cm]
    & \tiny{ $ \pm 1.9 $ } & \tiny{ $ \pm 1.5 $ } & \tiny{ $ \pm 1.0 $ } & \tiny{ $ \pm 4.1 $ } & \tiny{ $ \pm 4.8 $ } & \tiny{ $ \pm 3.6 $ } \\
    \textit{usedAs} & 49.2 & 36.5 & 41.8 & 30.1 & 19.1 & 23.2 & 0.0 & 0.0 & 0.0 \\[-0.15cm]
    & \tiny{ $ \pm 2.8 $ } & \tiny{ $ \pm 2.2 $ } & \tiny{ $ \pm 1.9 $ } & \tiny{ $ \pm 4.6 $ } & \tiny{ $ \pm 3.0 $ } & \tiny{ $ \pm 3.4 $ } \\
    \textit{conditionProperty} & 65.3 & 81.2 & 72.3 & 29.0 & 27.4 & 27.9 & 0.0 & 0.0 & 0.0 \\[-0.15cm]
    & \tiny{ $ \pm 2.1 $ } & \tiny{ $ \pm 2.5 $ } & \tiny{ $ \pm 1.0 $ } & \tiny{ $ \pm 3.6 $ } & \tiny{ $ \pm 2.6 $ } & \tiny{ $ \pm 1.8 $ } \\
    \textit{conditionPropertyValue} & 51.3 & 82.4 & 63.2 & 26.7 & 42.5 & 32.7 & 27.2 & 100.0 & 42.7 \\[-0.15cm]
    & \tiny{ $ \pm 3.0 $ } & \tiny{ $ \pm 3.3 $ } & \tiny{ $ \pm 2.5 $ } & \tiny{ $ \pm 2.8 $ } & \tiny{ $ \pm 2.7 $ } & \tiny{ $ \pm 2.4 $ } \\
    \textit{conditionSampleFeatures} & 60.7 & 72.3 & 66.0 & 34.4 & 28.4 & 31.0 & 70.4 & 42.8 & 53.2 \\[-0.15cm]
    & \tiny{ $ \pm 2.8 $ } & \tiny{ $ \pm 1.3 $ } & \tiny{ $ \pm 2.1 $ } & \tiny{ $ \pm 4.6 $ } & \tiny{ $ \pm 3.1 $ } & \tiny{ $ \pm 3.3 $ } \\
    \textit{usesTechnique} & 87.1 & 95.2 & 91.0 & 44.7 & 37.9 & 41.0 & 81.8 & 100.0 & 90.0 \\[-0.15cm]
    & \tiny{ $ \pm 1.1 $ } & \tiny{ $ \pm 1.3 $ } & \tiny{ $ \pm 0.9 $ } & \tiny{ $ \pm 1.6 $ } & \tiny{ $ \pm 2.7 $ } & \tiny{ $ \pm 1.8 $ } \\
    \textit{conditionEnvironment} & 40.4 & 37.9 & 39.0 & 32.2 & 27.0 & 29.2 & 0.0 & 0.0 & 0.0 \\[-0.15cm]
    & \tiny{ $ \pm 2.6 $ } & \tiny{ $ \pm 4.1 $ } & \tiny{ $ \pm 3.0 $ } & \tiny{ $ \pm 4.1 $ } & \tiny{ $ \pm 3.8 $ } & \tiny{ $ \pm 3.1 $ } \\
    \textit{propertyValue} & 78.7 & 86.8 & 82.5 & 40.9 & 46.7 & 43.4 & 0.0 & 0.0 & 0.0 \\[-0.15cm]
    & \tiny{ $ \pm 3.1 $ } & \tiny{ $ \pm 2.7 $ } & \tiny{ $ \pm 2.1 $ } & \tiny{ $ \pm 2.6 $ } & \tiny{ $ \pm 5.1 $ } & \tiny{ $ \pm 2.0 $ } \\
    \textit{usedIn} & 42.5 & 35.6 & 37.9 & 18.6 & 14.2 & 15.9 & 0.0 & 0.0 & 0.0 \\[-0.15cm]
    & \tiny{ $ \pm 4.9 $ } & \tiny{ $ \pm 7.6 $ } & \tiny{ $ \pm 3.7 $ } & \tiny{ $ \pm 7.1 $ } & \tiny{ $ \pm 3.6 $ } & \tiny{ $ \pm 4.8 $ } \\
    \textit{conditionInstrument} & 93.4 & 91.1 & 92.2 & 38.6 & 36.6 & 37.5 & 90.4 & 100.0 & 94.9 \\[-0.15cm]
    & \tiny{ $ \pm 0.1 $ } & \tiny{ $ \pm 1.6 $ } & \tiny{ $ \pm 0.9 $ } & \tiny{ $ \pm 1.9 $ } & \tiny{ $ \pm 3.7 $ } & \tiny{ $ \pm 2.5 $ } \\
    \textit{dopedBy} & 26.7 & 20.0 & 22.7 & 20.0 & 6.7 & 10.0 & 0.0 & 0.0 & 0.0 \\[-0.15cm]
    & \tiny{ $ \pm 22.6 $ } & \tiny{ $ \pm 16.3 $ } & \tiny{ $ \pm 18.7 $ } & \tiny{ $ \pm 40.0 $ } & \tiny{ $ \pm 13.3 $ } & \tiny{ $ \pm 20.0 $ } \\
    \textit{takenFrom} & 75.3 & 96.9 & 84.7 & 67.8 & 61.5 & 64.0 & 46.4 & 100.0 & 63.4 \\[-0.15cm]
    & \tiny{ $ \pm 5.1 $ } & \tiny{ $ \pm 3.8 $ } & \tiny{ $ \pm 3.6 $ } & \tiny{ $ \pm 7.8 $ } & \tiny{ $ \pm 6.9 $ } & \tiny{ $ \pm 4.6 $ } \\
    \textit{usedTogether} & 9.6 & 2.5 & 4.0 & 9.6 & 2.4 & 3.8 & 0.0 & 0.0 & 0.0 \\[-0.15cm]
    & \tiny{ $ \pm 3.5 $ } & \tiny{ $ \pm 1.1 $ } & \tiny{ $ \pm 1.7 $ } & \tiny{ $ \pm 4.0 $ } & \tiny{ $ \pm 1.1 $ } & \tiny{ $ \pm 1.7 $ } \\
    \midrule
    Macro-avg. & 60.2 & 65.5 & 62.0 & 35.2 & 32.0 & 32.8 & 24.4 & 36.1 & 27.4\\[-0.15cm]
    & \tiny{ $ \pm 0.8 $ } & \tiny{ $ \pm 1.6 $ } & \tiny{ $ \pm 1.0 $ } & \tiny{ $ \pm 2.7 $ } & \tiny{ $ \pm 1.6 $ } & \tiny{ $ \pm 1.2 $ } \\
    Micro-avg. & 66.8 & 68.4 & 67.6 & 38.6 & 34.7 & 36.5 & 50.5 & 29.5 & 37.2\\[-0.15cm]
    & \tiny{ $ \pm 1.7 $ } & \tiny{ $ \pm 0.4 $ } & \tiny{ $ \pm 1.0 $ } & \tiny{ $ \pm 2.3 $ } & \tiny{ $ \pm 1.3 $ } & \tiny{ $ \pm 0.9 $ } \\
    \bottomrule
    \end{tabular}
    \caption{Per-label scores (\corpusNameShort test set) for \textbf{relation extraction} using MatSciBERT. Majority baseline is computed on gold entities.}
    \label{tab:all_relation_results_test}
\end{table*}

\begin{table*}[htb]
    \centering
    \footnotesize
    \begin{tabular}{lccc|ccc|ccc}
    \toprule
     & \multicolumn{3}{c}{\textbf{\corpusNameShort only}} & \multicolumn{3}{c}{\textbf{\corpusNameShort + SOFC-Exp}} &  \multicolumn{3}{c}{\textbf{\corpusNameShort + MSPT}}\\
    \textbf{Label} & \textbf{P} & \textbf{R} & \textbf{F1} & \textbf{P} & \textbf{R} & \textbf{F1} & \textbf{P} & \textbf{R} & \textbf{F1} \\
     \midrule
    \textit{hasForm} & 74.0 & 78.5 & 76.1 & 78.0 & 78.8 & 78.3 & 74.2 & 79.6 & 76.8 \\[-0.15cm]
 & \tiny{ $ \pm 1.8 $ } & \tiny{ $ \pm 1.1 $ } & \tiny{ $ \pm 0.5 $ } & \tiny{ $ \pm 2.9 $ } & \tiny{ $ \pm 1.8 $ } & \tiny{ $ \pm 0.9 $ } & \tiny{ $ \pm 0.7 $ } & \tiny{ $ \pm 1.3 $ } & \tiny{ $ \pm 0.9 $ } \\
\textit{measuresProperty} & 80.2 & 86.3 & 83.1 & 79.0 & 87.1 & 82.8 & 78.8 & 86.6 & 82.5 \\[-0.15cm]
 & \tiny{ $ \pm 1.1 $ } & \tiny{ $ \pm 2.2 $ } & \tiny{ $ \pm 0.8 $ } & \tiny{ $ \pm 1.1 $ } & \tiny{ $ \pm 0.9 $ } & \tiny{ $ \pm 0.4 $ } & \tiny{ $ \pm 1.2 $ } & \tiny{ $ \pm 1.3 $ } & \tiny{ $ \pm 0.3 $ } \\
\textit{measuresPropertyValue} & 68.6 & 80.0 & 73.8 & 68.9 & 82.7 & 75.1 & 70.3 & 82.1 & 75.7 \\[-0.15cm]
 & \tiny{ $ \pm 1.9 $ } & \tiny{ $ \pm 1.5 $ } & \tiny{ $ \pm 1.0 $ } & \tiny{ $ \pm 3.5 $ } & \tiny{ $ \pm 2.2 $ } & \tiny{ $ \pm 2.0 $ } & \tiny{ $ \pm 0.7 $ } & \tiny{ $ \pm 3.1 $ } & \tiny{ $ \pm 1.5 $ } \\
\textit{usedAs} & 49.2 & 36.5 & 41.8 & 47.9 & 35.7 & 40.8 & 51.9 & 37.5 & 43.5 \\[-0.15cm]
 & \tiny{ $ \pm 2.8 $ } & \tiny{ $ \pm 2.2 $ } & \tiny{ $ \pm 1.9 $ } & \tiny{ $ \pm 2.6 $ } & \tiny{ $ \pm 2.0 $ } & \tiny{ $ \pm 1.3 $ } & \tiny{ $ \pm 1.3 $ } & \tiny{ $ \pm 1.8 $ } & \tiny{ $ \pm 1.4 $ } \\
\textit{conditionProperty} & 65.3 & 81.2 & 72.3 & 67.4 & 82.1 & 73.9 & 66.1 & 80.3 & 72.5 \\[-0.15cm]
 & \tiny{ $ \pm 2.1 $ } & \tiny{ $ \pm 2.5 $ } & \tiny{ $ \pm 1.0 $ } & \tiny{ $ \pm 2.5 $ } & \tiny{ $ \pm 2.5 $ } & \tiny{ $ \pm 0.8 $ } & \tiny{ $ \pm 1.9 $ } & \tiny{ $ \pm 2.2 $ } & \tiny{ $ \pm 1.6 $ } \\
\textit{conditionPropertyValue} & 51.3 & 82.4 & 63.2 & 53.7 & 77.8 & 63.3 & 51.1 & 78.5 & 61.8 \\[-0.15cm]
 & \tiny{ $ \pm 3.0 $ } & \tiny{ $ \pm 3.3 $ } & \tiny{ $ \pm 2.5 $ } & \tiny{ $ \pm 4.2 $ } & \tiny{ $ \pm 4.3 $ } & \tiny{ $ \pm 2.0 $ } & \tiny{ $ \pm 2.6 $ } & \tiny{ $ \pm 3.1 $ } & \tiny{ $ \pm 2.1 $ } \\
\textit{conditionSampleFeatures} & 60.7 & 72.3 & 66.0 & 61.1 & 71.4 & 65.7 & 62.8 & 72.2 & 67.1 \\[-0.15cm]
 & \tiny{ $ \pm 2.8 $ } & \tiny{ $ \pm 1.3 $ } & \tiny{ $ \pm 2.1 $ } & \tiny{ $ \pm 3.7 $ } & \tiny{ $ \pm 3.8 $ } & \tiny{ $ \pm 1.4 $ } & \tiny{ $ \pm 3.4 $ } & \tiny{ $ \pm 2.5 $ } & \tiny{ $ \pm 2.6 $ } \\
\textit{usesTechnique} & 87.1 & 95.2 & 91.0 & 85.6 & 97.1 & 91.0 & 87.2 & 95.7 & 91.2 \\[-0.15cm]
 & \tiny{ $ \pm 1.1 $ } & \tiny{ $ \pm 1.3 $ } & \tiny{ $ \pm 0.9 $ } & \tiny{ $ \pm 0.8 $ } & \tiny{ $ \pm 1.0 $ } & \tiny{ $ \pm 0.3 $ } & \tiny{ $ \pm 1.1 $ } & \tiny{ $ \pm 0.8 $ } & \tiny{ $ \pm 0.6 $ } \\
\textit{conditionEnvironment} & 40.4 & 37.9 & 39.0 & 46.3 & 46.2 & 46.0 & 47.2 & 49.8 & 48.3 \\[-0.15cm]
 & \tiny{ $ \pm 2.6 $ } & \tiny{ $ \pm 4.1 $ } & \tiny{ $ \pm 3.0 $ } & \tiny{ $ \pm 3.5 $ } & \tiny{ $ \pm 4.6 $ } & \tiny{ $ \pm 2.7 $ } & \tiny{ $ \pm 6.3 $ } & \tiny{ $ \pm 4.7 $ } & \tiny{ $ \pm 5.2 $ } \\
\textit{propertyValue} & 78.7 & 86.8 & 82.5 & 76.2 & 85.5 & 80.5 & 81.4 & 88.7 & 84.8 \\[-0.15cm]
 & \tiny{ $ \pm 3.1 $ } & \tiny{ $ \pm 2.7 $ } & \tiny{ $ \pm 2.1 $ } & \tiny{ $ \pm 2.7 $ } & \tiny{ $ \pm 1.7 $ } & \tiny{ $ \pm 1.2 $ } & \tiny{ $ \pm 1.7 $ } & \tiny{ $ \pm 1.7 $ } & \tiny{ $ \pm 1.1 $ } \\
\textit{usedIn} & 42.5 & 35.6 & 37.9 & 45.5 & 47.6 & 45.9 & 41.8 & 40.4 & 40.5 \\[-0.15cm]
 & \tiny{ $ \pm 4.9 $ } & \tiny{ $ \pm 7.6 $ } & \tiny{ $ \pm 3.7 $ } & \tiny{ $ \pm 7.0 $ } & \tiny{ $ \pm 5.7 $ } & \tiny{ $ \pm 3.7 $ } & \tiny{ $ \pm 4.9 $ } & \tiny{ $ \pm 5.5 $ } & \tiny{ $ \pm 1.4 $ } \\
\textit{conditionInstrument} & 93.4 & 91.1 & 92.2 & 93.6 & 93.2 & 93.4 & 93.5 & 91.9 & 92.7 \\[-0.15cm]
 & \tiny{ $ \pm 0.1 $ } & \tiny{ $ \pm 1.6 $ } & \tiny{ $ \pm 0.9 $ } & \tiny{ $ \pm 0.1 $ } & \tiny{ $ \pm 0.9 $ } & \tiny{ $ \pm 0.5 $ } & \tiny{ $ \pm 0.1 $ } & \tiny{ $ \pm 1.6 $ } & \tiny{ $ \pm 0.9 $ } \\
\textit{dopedBy} & 26.7 & 20.0 & 22.7 & 0.0 & 0.0 & 0.0 & 28.3 & 26.7 & 27.0 \\[-0.15cm]
 & \tiny{ $ \pm 22.6 $ } & \tiny{ $ \pm 16.3 $ } & \tiny{ $ \pm 18.7 $ } & \tiny{ $ \pm 0.0 $ } & \tiny{ $ \pm 0.0 $ } & \tiny{ $ \pm 0.0 $ } & \tiny{ $ \pm 16.3 $ } & \tiny{ $ \pm 13.3 $ } & \tiny{ $ \pm 14.0 $ } \\
\textit{takenFrom} & 75.3 & 96.9 & 84.7 & 61.7 & 98.5 & 75.6 & 77.0 & 98.5 & 86.2 \\[-0.15cm]
 & \tiny{ $ \pm 5.1 $ } & \tiny{ $ \pm 3.8 $ } & \tiny{ $ \pm 3.6 $ } & \tiny{ $ \pm 7.5 $ } & \tiny{ $ \pm 3.1 $ } & \tiny{ $ \pm 5.7 $ } & \tiny{ $ \pm 8.6 $ } & \tiny{ $ \pm 3.1 $ } & \tiny{ $ \pm 6.2 $ } \\
\textit{usedTogether} & 9.6 & 2.5 & 4.0 & 8.8 & 2.2 & 3.5 & 12.0 & 3.6 & 5.4 \\[-0.15cm]
 & \tiny{ $ \pm 3.5 $ } & \tiny{ $ \pm 1.1 $ } & \tiny{ $ \pm 1.7 $ } & \tiny{ $ \pm 3.3 $ } & \tiny{ $ \pm 0.8 $ } & \tiny{ $ \pm 1.2 $ } & \tiny{ $ \pm 7.2 $ } & \tiny{ $ \pm 3.4 $ } & \tiny{ $ \pm 4.8 $ } \\
 \midrule
Macro-avg. & 60.2 & 65.5 & 62.0 & 58.2 & 65.7 & 61.1 & 61.6 & 67.5 & 63.7 \\[-0.15cm]
 & \tiny{ $ \pm 0.8 $ } & \tiny{ $ \pm 1.6 $ } & \tiny{ $ \pm 1.0 $ } & \tiny{ $ \pm 1.5 $ } & \tiny{ $ \pm 1.1 $ } & \tiny{ $ \pm 0.7 $ } & \tiny{ $ \pm 0.9 $ } & \tiny{ $ \pm 1.0 $ } & \tiny{ $ \pm 0.7 $ } \\
Micro-avg. & 66.8 & 68.4 & 67.6 & 67.4 & 68.9 & 68.1 & 68.0 & 69.5 & 68.7 \\[-0.15cm]
 & \tiny{ $ \pm 1.7 $ } & \tiny{ $ \pm 0.4 $ } & \tiny{ $ \pm 1.0 $ } & \tiny{ $ \pm 1.9 $ } & \tiny{ $ \pm 1.2 $ } & \tiny{ $ \pm 0.7 $ } & \tiny{ $ \pm 1.1 $ } & \tiny{ $ \pm 0.5 $ } & \tiny{ $ \pm 0.7 $ } \\
    \bottomrule
    \end{tabular}
    \caption{Per-label scores (\corpusNameShort test set, gold entities) for \textbf{multi-task relation extraction}.}
    \label{tab:all_relation_results_mtl}
\end{table*}

\end{document}